\documentclass{article} 
\usepackage{iclr2027_conference,times}

\usepackage{amsmath,amsfonts,bm}

\def\eqref#1{equation~\ref{#1}}

\def\1{\bm{1}}

\DeclareMathAlphabet{\mathsfit}{\encodingdefault}{\sfdefault}{m}{sl}
\SetMathAlphabet{\mathsfit}{bold}{\encodingdefault}{\sfdefault}{bx}{n}

\usepackage{graphicx}
\usepackage{hyperref}
\usepackage{url}
\usepackage{array}
\usepackage{tabularx}
\usepackage{wrapfig}
\usepackage{booktabs}
\usepackage{makecell}
\usepackage{xcolor}
\usepackage{caption}
\usepackage{multirow}
\usepackage{amssymb}
\newcommand{\xmark}{\textcolor{red}{\ensuremath{\bm{\times}}}}
\newcommand{\cmark}{\textcolor{green!45!black}{\ensuremath{\bm{\checkmark}}}}
\title{Across Viewpoints and Time: Trajectory-Generalizable Physical Attacks on World Models and Video Understanding Models}

\author{
Song Xia\textsuperscript{1},
Meiewen Ding\textsuperscript{1},
Chenqi Kong\textsuperscript{2},
Wenhan Yang\textsuperscript{3},
Xudong Jiang\textsuperscript{1}\\
\textsuperscript{1}Nanyang Technological University,
\textsuperscript{2}National University of Singapore,
\textsuperscript{3}Pengcheng Lab\\
\texttt{\{Xias0002,EXDJiang\}@ntu.edu.sg}
}

\begin{document}

\maketitle
\begin{abstract}
World models and video understanding models are increasingly deployed in dynamic visual environments, yet their robustness to physical adversarial threats remains largely underexplored.
We study whether a single physical adversarial texture can mislead the victim model across different real-world scenarios where observing viewpoints vary along uncontrolled object--camera trajectories.
We identify that heterogeneous video semantics and trajectories induce poorly aligned gradients for optimizing the adversarial texture.
To address that, we propose the Trajectory-aware adversarial Alignment and Refinement Attack (TARA), which incorporates trajectory-aware optimization to capture the evolving geometric variations of a printed texture under real-world scenarios.
To improve cross-instance optimization compatibility, TARA incorporates trajectory-aware contrastive alignment to encourage attacked representations from diverse video--trajectory instances to approach a shared adversarial region and then performs trajectory-aware task refinement to translate the aligned representation into task-level prediction errors.
For reproducible evaluation, we develop a geometry-grounded tracking pipeline that provides physically plausible adversarial regions with temporally consistent trajectories on EPIC-KITCHENS-100 and Something-Something V2.
We further collect real-world print-and-recapture videos to validate attack effectiveness under physical deployment conditions.
Together, the annotated benchmarks and real-world captured data form a reproducible testbed for future evaluation.
Notably, our results reveal that a single printable adversarial texture can persistently disrupt world models and video understanding models under physical conditions, highlighting the fundamental security challenge.

\end{abstract}
\vspace{-3mm}
\section{Introduction}
\vspace{-2mm}
World models~\citep{assran2025v,wu2024ivideogpt,zhou2025dino,zhang2026dila,zhang2026structure} and video foundation models~\citep{tong2022videomae,wang2023videomae,wang2024internvideo2,zhao2024videoprism} are advancing visual understanding and prediction in dynamic environments.
Their deployment in real-world systems raises concerns about vulnerability to physical adversarial inputs.
Adversaries in video understanding and action anticipation have been extensively studied in the digital domain~\citep{wei2019sparse,wei2022cross,xia2024transferable,ding2026transferable,lu2026pretrain}.
However, attacks applied directly to model inputs do not establish whether similar vulnerabilities exist under physical deployment ~\citep{brown2017adversarial,athalye2018synthesizing,eykholt2018robust}.
In physical settings, an attacker can design and attach an adversarial texture but cannot control how it is subsequently observed.
As object--camera trajectories evolve, the attached texture undergoes substantial variations in viewpoint, scale, and visibility, potentially limiting its attack effectiveness.
This raises a critical concern about the practical vulnerability of video understanding and anticipation models to physical adversarial attacks under uncontrolled observation conditions in real-world deployments.

Previous physical attacks consider constraints such as fabrication, camera acquisition, and geometric variation~\citep{brown2017adversarial,athalye2018synthesizing,eykholt2018robust}, with subsequent work extending to video recognition, tracking, and embodied systems~\citep{pony2021over,du2022motional,long2024papmot,chen2024diffusion,li2025uv,wang2025exploring}.
While these studies increasingly account for variations in how adversarial patterns are observed, their effectiveness is often assessed under predefined conditions, which may not fully reflect the real-world variation~\citep{hingun2023reap}.
Specifically, the ability of a single physical adversarial texture to generalize jointly across heterogeneous videos and unseen object--camera trajectories remains insufficiently explored.

\begin{figure}
    \centering
    \includegraphics[width=0.88\linewidth]{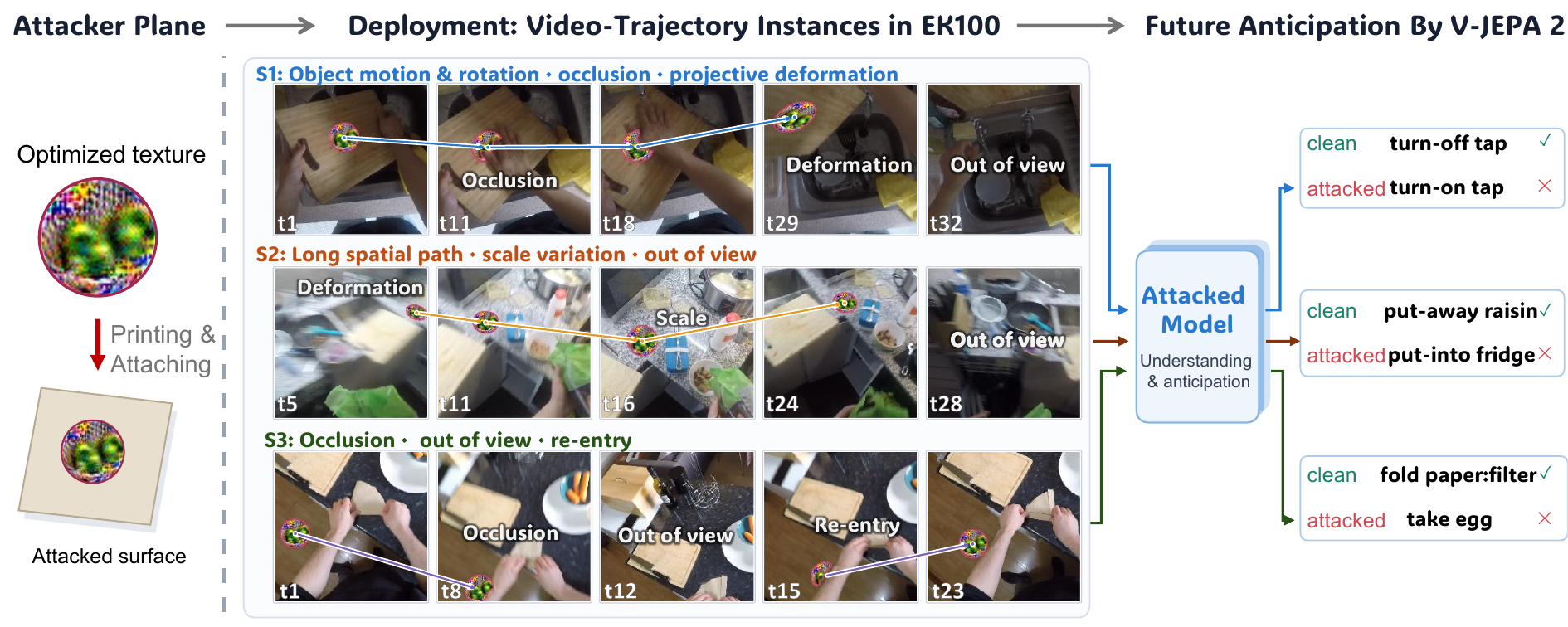}
    \vspace{-2mm}
    \caption{Illustration of physical adversarial texture on action anticipation under various trajectories.}
    \label{fig:teaser}
    \vspace{-5mm}
\end{figure}

To investigate this setting, as shown in \figureautorefname~\ref{fig:teaser}, we consider a physical attack where a single printable texture is optimized before deployment and must remain effective across unseen object--camera trajectories beyond the attacker's control.
However, jointly optimizing the texture across heterogeneous videos and observation trajectories can induce conflicting gradients, limiting the attack effectiveness.
To solve that, we propose a Trajectory-aware adversarial Alignment and Refinement Attack(TARA), which optimizes a single texture across diverse video--trajectory instances to account for evolving physical observation conditions.
TARA first employs trajectory-aware contrastive alignment to encourage attacked representations across heterogeneous instances to approach a shared adversarial region, improving cross-instance compatibility.
The resulting texture is then refined along each video's trajectory to induce task-level prediction errors.
Notably, evaluating physical attacks on video models requires temporally coherent surface trajectories that capture how a fixed texture is observed over time, which existing image-level benchmarks~\citep{braunegg2020apricot,hingun2023reap} do not provide. 
To support that, we develop a geometry-grounded tracking pipeline that identifies feasible attachment regions and generates temporally consistent homographies and visibility masks for Something-Something V2~\citep{goyal2017something} and EPIC-KITCHENS-100~\citep{damen2022rescaling} datasets.
Meanwhile, we further collect real-world print-and-recapture videos, in which learned textures are attached to real objects and evaluated under physical conditions.
We curate surface masks and trajectory annotations for both simulated and real-world captured videos, establishing a reproducible evaluation benchmark for future research.

We evaluate TARA across multiple world models and video understanding models on recognition and future prediction tasks.
Our results demonstrate that TARA maintains its attack effectiveness across unseen video--trajectory instances and real-world print-and-recapture scenarios, highlighting the vulnerabilities of those models in practical deployment.
Overall, our contributions are:
\begin{itemize}
\vspace{-1mm}
    \item We investigate a practical physical attack for world models and video understanding models, where a single printable texture must generalize across heterogeneous videos and object--camera trajectories after deployment.
\vspace{-0.5mm}
    \item We identify conflicting optimization gradients across heterogeneous video--trajectory instances as a key challenge in learning a shared texture. To address this, we propose TARA, which integrates trajectory-aware optimization with cross-instance contrastive alignment and task-level refinement to improve effectiveness under dynamic physical conditions.
\vspace{-0.5mm}
    \item We establish a geometry-grounded benchmark spanning Something-Something V2, EPIC-KITCHENS-100, and real-world print-and-recapture videos, providing temporally consistent trajectories with homographies and visibility masks for evaluating physical attacks.
\end{itemize}

\vspace{-2mm}
\section{Preliminaries}
\label{sec:pre}
\vspace{-1mm}
\textbf{Attack setting.}
Denote $F_{\theta}$ as the attacked model and $\bm{X}_i=\{\bm{x}_{it}\}_{t=1}^{T_i}$ as the video clip with ground truth $y_i$.
%
%
We assume a white-box adversary who optimizes a single printable adversarial texture $\bm{\delta}$ to disrupt model predictions across heterogeneous videos and observation trajectories.
Practically, we assume the texture $\bm{\delta}$ is attached to a fixed object surface, while its image-space appearance evolves with the underlying object--camera trajectory and scene visibility, which are beyond the adversary's control after deployment

\begin{wrapfigure}{r}{0.46\columnwidth}
    \centering
    \vspace{-6mm}
    \includegraphics[width=\linewidth]{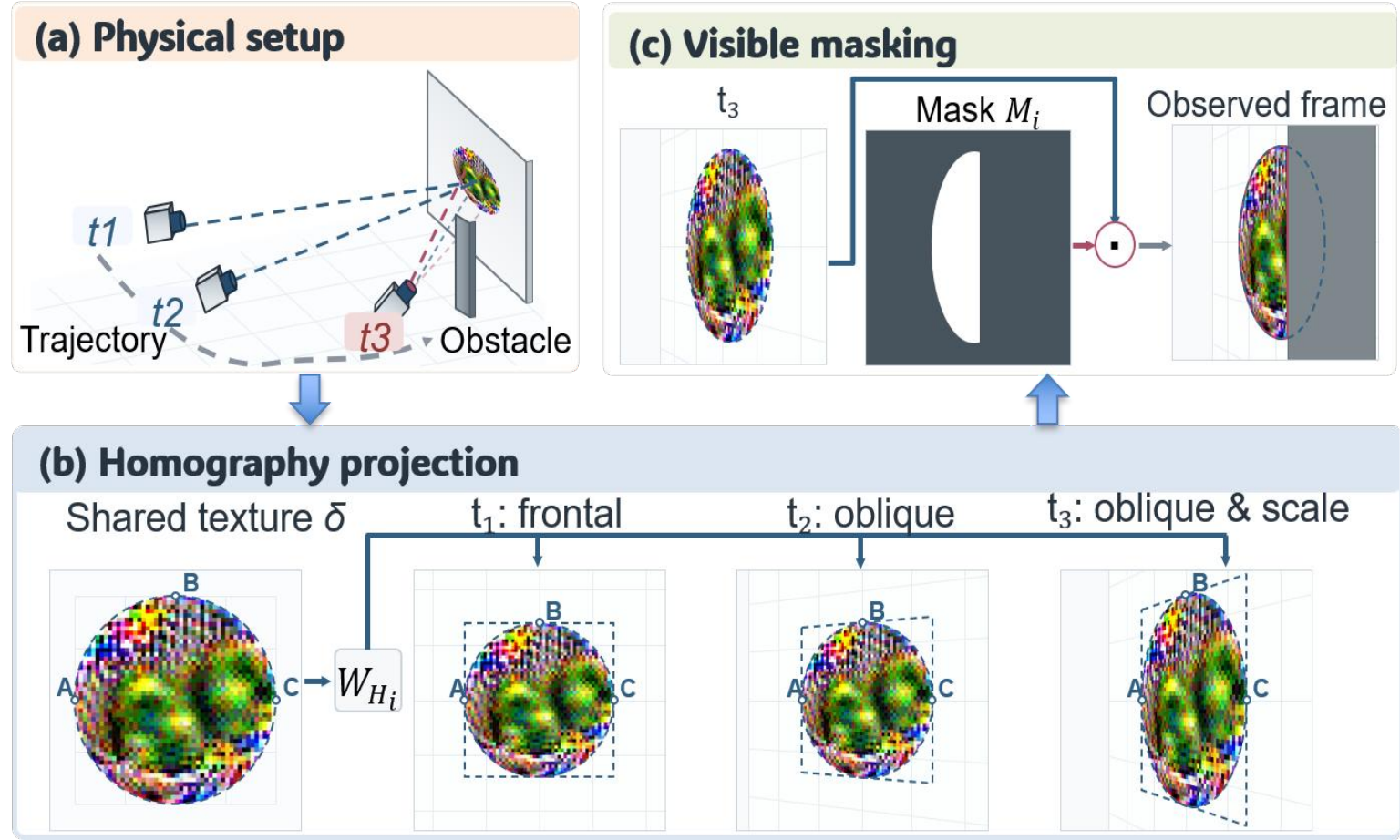}
    \caption{Illustration of the observation trajectory of adversarial texture.}
    \label{fig:proj_illu}
    \vspace{-3mm}
\end{wrapfigure}

\textbf{Homography and visibility mask.} For frame $t$ of video $i$, we characterize the observation state of the attached texture by a homography $\bm{H}_{it}$ and a visibility mask $\bm{M}_{it}$.
As illustrated in \figureautorefname~\ref{fig:proj_illu}, $\bm{H}_{it}$ maps the adversarial texture from its local coordinates to the image plane, accounting for changes in position, scale, orientation, and perspective distortion.
The visibility mask $\bm{M}_{it}\in[0,1]^{h\times w}$ specifies the observable pixels of the projected texture after occlusion.
We define the \emph{observation trajectory} of video $i$ as $\gamma_i=\{(\bm{H}_{it},\bm{M}_{it})\}_{t=1}^{T_i}$, capturing the temporally coupled evolution of texture homography and visibility.
Let $\mathcal{W}_{\bm{H}_{it}}$ be the differentiable projective matrix.
Given the adversarial texture $\bm{\delta}$, the attacked frame--trajectory instance $\tilde{\bm{x}}_{it}^{\,\bm{\delta}}$ and video--trajectory instance $\tilde{\bm{X}}_i(\bm{\delta};\gamma_i)$ are constructed as:
\begin{equation}
\tilde{\bm{x}}_{it}^{\,\bm{\delta}}
=
(1-\bm{M}_{it})\odot\bm{x}_{it}
+
\bm{M}_{it}\odot\mathcal{W}_{\bm{H}_{it}}(\bm{\delta}),
\qquad
\tilde{\bm{X}}_i(\bm{\delta};\gamma_i)
=
\{\tilde{\bm{x}}_{it}^{\,\bm{\delta}}\}_{t=1}^{T_i}.
\label{eq:gen_of_adv_video}
\end{equation}

\vspace{-3mm}
\section{Methodology}
\vspace{-2mm}
\subsection{Trajectory-Aware Adversarial Alignment and Refinement Attack}
\vspace{-1mm}
\label{sec:trajectory_contrastive_alignment}
To account for the evolving physical observation conditions, we optimize a shared adversarial texture across diverse video--trajectory instances, where each instance captures the temporally coupled variations in texture homography and visibility.
Given $N$ training instances, a direct optimization aims to find the $\bm{\delta}$ that maximize a task-specific objective $\mathcal{L}_{\mathrm{dir}}$ across all instances:
\begin{equation}
\bm{\delta}^{\star}
=\arg\max_{\bm{\delta}}\frac{1}{N}
\sum_{i=1}^{N}\mathcal{L}_{\mathrm{dir}}
\left(F_{\theta}\!\left(\tilde{\bm{X}}_i(\bm{\delta};\gamma_i)\right),y_i\right).
\label{eq:direct_task_objective}
\end{equation}
\begin{wrapfigure}{r}{0.44\columnwidth}
    \centering
    \vspace{-4mm}
    \includegraphics[width=\linewidth]{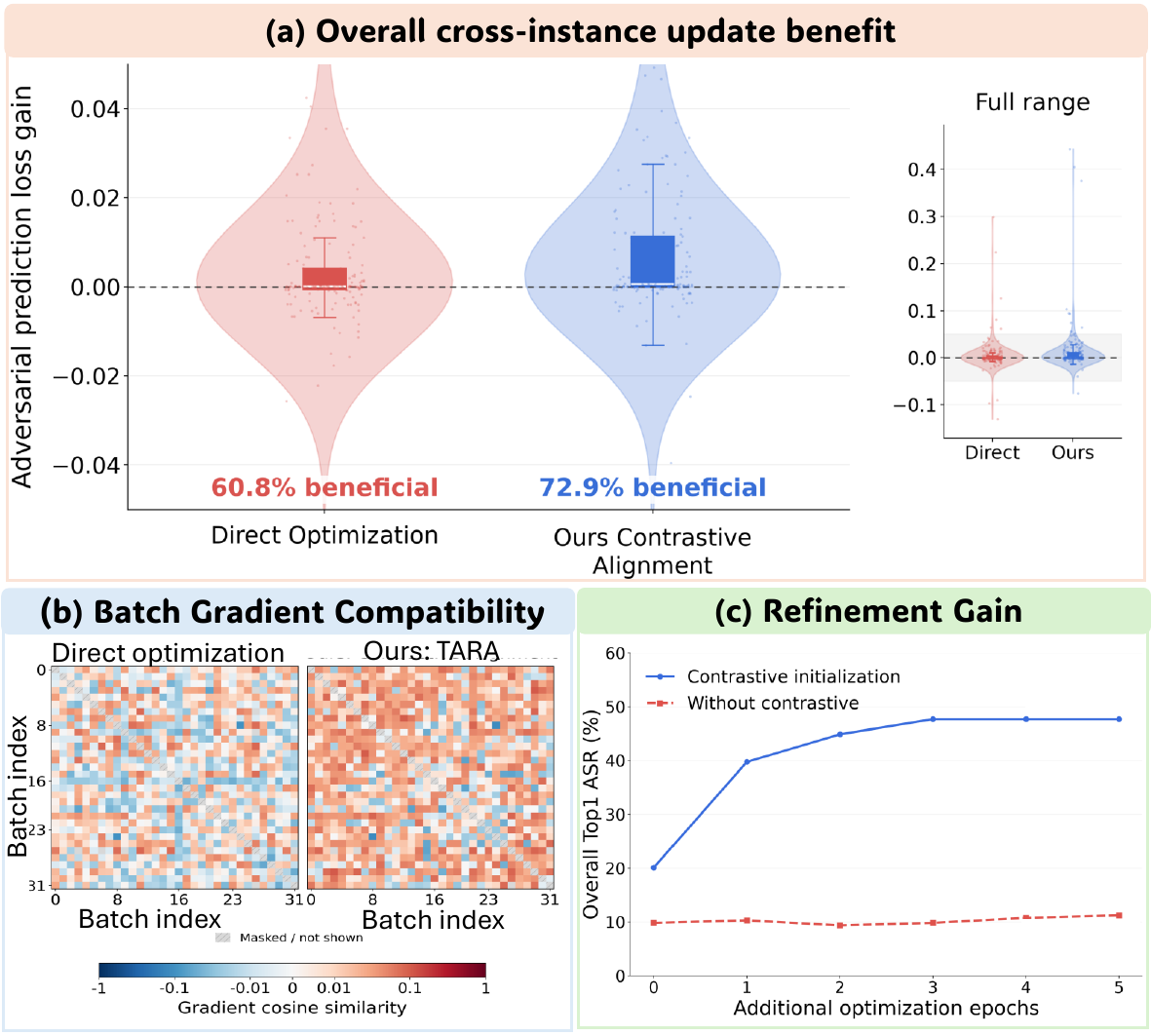}
    \caption{
    Analysis of trajectory-aware contrastive alignment.
    }
    \label{fig:alignment_analysis}
    \vspace{-5mm}
\end{wrapfigure}
However, jointly optimizing the texture across heterogeneous video--trajectory instances can produce conflicting gradients, limiting the effectiveness of shared texture updates. 
Figure~\ref{fig:alignment_analysis}(a) and (b) report the average cross-instance update benefits and pairwise gradient similarities, reporting statistics averaged over all training clips and optimization iterations.
The results reveal limited cross-instance update benefits and substantial gradient disagreement under direct optimization.
These motivate us to propose Trajectory-aware adversarial Alignment and Refinement Attack (TARA), which first aligns adversarial representations across video--trajectory instances to improve optimization compatibility, followed by task refinement to induce prediction errors.
%

\begin{figure}
    \centering
    \includegraphics[width=1\linewidth]{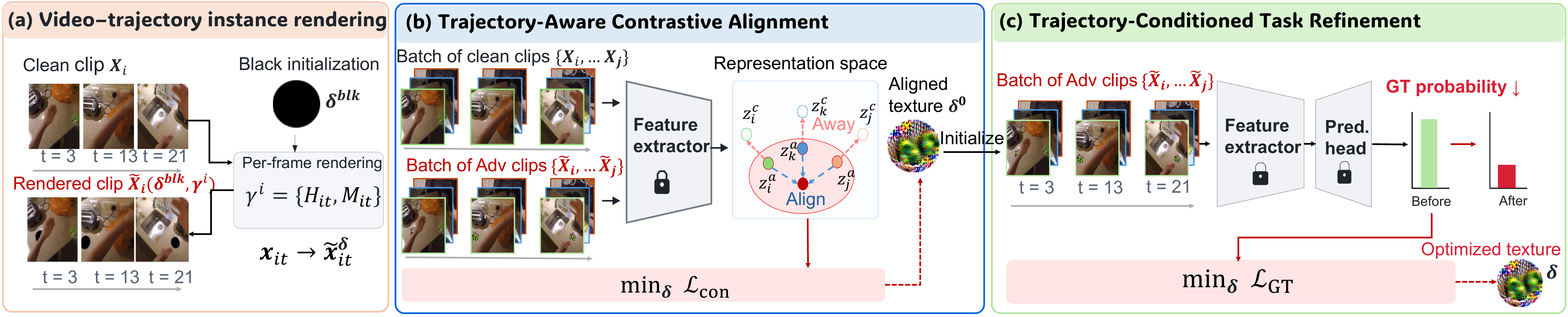}
    \vspace{-4mm}
    \caption{Overall framework of TARA. A black texture is rendered along the video-trajectory instance and optimized through trajectory-aware contrastive alignment and task refinement.}
    \label{fig:framework}
    \vspace{-5mm}
\end{figure}

\textbf{Trajectory-Aware Contrastive Alignment.}
As shown in \figureautorefname~\ref{fig:framework}(a) and (b), we pair each training video $\bm{X}_i$ with its observation trajectory $\gamma_i$ to construct the attacked video--trajectory instance $\tilde{\bm{X}}_i(\bm{\delta};\gamma_i)$.
We first extract its attacked representation $\bm{z}_i^a$ and the corresponding clean representation $\bm{z}_i^c$.
To encourage compatible adversarial representations across heterogeneous video--trajectory instances, for a batch of $B$ instances, we minimize the following contrastive loss $\mathcal{L}_{\mathrm{con}}$:
\begin{equation}
\mathcal{L}_{\mathrm{con}}
=
\frac{1}{B(B-1)}\sum_{i\neq j}\operatorname{softplus}
\left(
\frac{s(\bm{z}^{a}_{i},\bm{z}^{c}_{i})-s(\bm{z}^{a}_{i},\bm{z}^{a}_{j})}{\tau}
\right),
\label{eq:trajectory_contrastive}
\end{equation}
where $s(\cdot,\cdot)$ denotes cosine similarity and $\tau$ is a temperature parameter.
Minimizing \equationautorefname~\ref{eq:trajectory_contrastive} encourages attacked representations from different video--trajectory instances to approach a shared adversarial region while remaining separated from their clean representations.
Optimization of $\mathcal{L}_{\mathrm{con}}$ yields the aligned adversarial texture $\bm{\delta}^{(0)}$, which serves as the initialization for task refinement.
Importantly, we find this alignment greatly improves cross-instance optimization compatibility.
As indicated in \figureautorefname~\ref{fig:alignment_analysis} (a) and (b), the contrastive alignment improves both average cross-instance update benefit and batch gradient agreement.
Moreover, the resulting texture also provides a stronger initialization that brings great performance gain under the same refinement, shown in \figureautorefname~\ref{fig:alignment_analysis} (c).

\textbf{Trajectory-Aware Task Refinement.}
While contrastive alignment encourages shared adversarial representations across video--trajectory instances, it does not explicitly optimize task-level prediction errors.
We therefore refine the aligned texture $\bm{\delta}^{(0)}$ along each training video's observation trajectory by minimizing:
\begin{equation}
\mathcal{L}_{\mathrm{ref}}
=
\sum_{i=1}^{B} w_i
\log p_{\theta}\!\left(
y_i \mid
\tilde{\bm{X}}_i(\bm{\delta};\gamma_i)
\right),
\label{eq:trajectory_task_refinement}
\end{equation}
where $w_i$ is a trajectory-aware weight inversely related to the projected visible texture area, computed from the frame-wise homographies and visibility masks.
By assigning larger weights to instances with smaller observable texture regions, this weighting balances their contributions to shared texture optimization under heterogeneous observation conditions.
More details of the weighting scheme are provided in the Appendix~\ref{app:trajectory_weighting}.

\vspace{-1mm}
\subsection{Geometry-Grounded Trajectory Construction}
\vspace{-1mm}
To support trajectory-aware optimization and reproducible evaluation, we construct geometry-grounded texture trajectories on Something-Something V2 (SSV2)~\citep{goyal2017something} and EPIC-KITCHENS-100 (EK100)~\citep{damen2022rescaling}.
For each video clip, we manually identify a feasible texture attachment region on an approximately planar object surface.
The region remains fixed on the object, while its image-space projection evolves with the underlying object--camera motion.

\textbf{Adversarial texture homography and visibility.}
Starting from the selected attachment region in a reference frame, we sample anchor points within the region and track them across frames using TAPNext++~\citep{Jung_2026_CVPR,zholus2025tapnext}.
We estimate a reference-to-frame homography from the tracked correspondences using RANSAC~\citep{fischler1981random}, which is then composed with the fixed texture-to-reference mapping to obtain the texture-to-frame homography $\bm{H}_{it}$.
When point tracking becomes unreliable, we leverage dense correspondences from RoMa~\citep{edstedt2024roma} with nearby validated frames to recover geometrically consistent projections.
The estimated homography determines the projected texture region in each frame, from which we further estimate its visible portion.
Specifically, GroundingDINO~\citep{liu2024grounding} detects supporting surfaces and foreground occluders, such as hands and arms, while SAM2~\citep{ravi2025sam} converts the detections into pixel-level masks.
We combine the projected support with image boundaries and surface masks, and remove foreground-occluded pixels to obtain the visibility mask $\bm{M}_{it}$.
The final sequence $\gamma_i=\{(\bm{H}_{it},\bm{M}_{it})\}_{t=1}^{T_i}$ defines the surface-observation trajectory and is further manually reviewed for consistency, drift, and occlusion quality.
Further implementation and quality-control details are provided in the Appendix~\ref{app:surface_trajectory}.
\begin{table*}[t]
\centering
\caption{Statistics of the geometry-grounded trajectory benchmarks.}
\vspace{-3mm}
\label{tab:trajectory_stats}
\setlength{\tabcolsep}{5pt}
\renewcommand{\arraystretch}{1.08}
\resizebox{0.90\textwidth}{!}{
\begin{tabular}{llccccccc}
\toprule
\textbf{Dataset} & \textbf{Task} & \textbf{Clips} & \makecell{\textbf{Texture}\\\textbf{Train/Eval.}} & \makecell{\textbf{Annotated}\\\textbf{Frames}} & \makecell{\textbf{Homography}\\\textbf{Coverage}} & \makecell{\textbf{Mask}\\\textbf{Rate}} & \makecell{\textbf{Texture}\\\textbf{Area}} & \makecell{\textbf{Texture}\\\textbf{Motion}} \\
\midrule
EK100 & Anticipation & 605 & 356 / 249 & 19,251 & 73.94\% & 7.56\% & 1.95\% & 20.65\% \\
SSV2 & Recognition & 328 & 184 / 144 & 10,204 & 89.78\% & 9.70\% & 2.37\% & 7.66\% \\
Real-world & Recog. / Antic. & 348 & 0 / 348 & 8,645 & 81.27\% & 4.28\% & 2.97\% & 15.17\% \\
\bottomrule
\end{tabular}}
\end{table*}
\begin{figure*}[t]
\vspace{-4mm}
\centering
\includegraphics[width=0.92\textwidth]{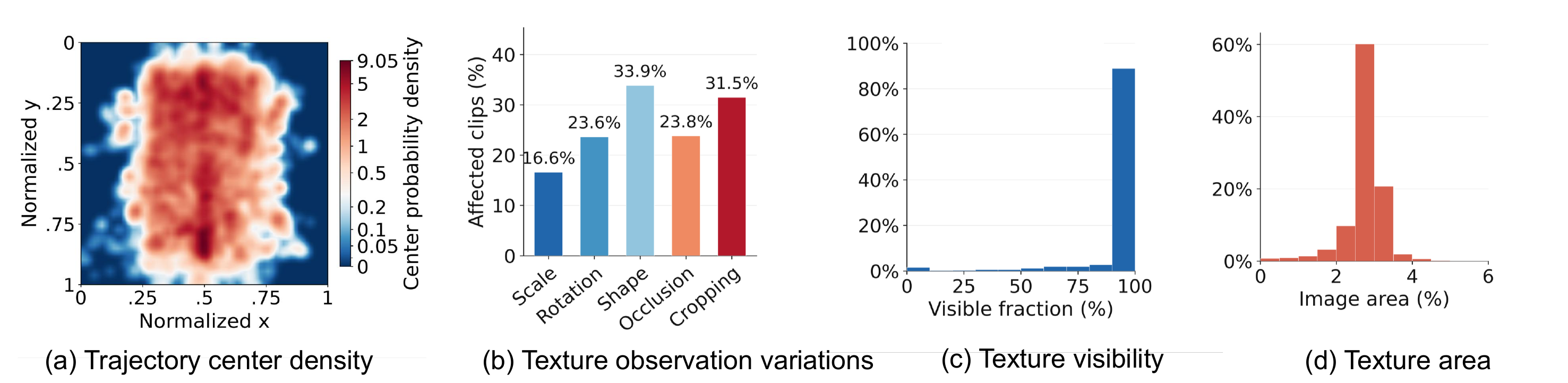}
\vspace{-2mm}
\caption{Distribution of surface-observation trajectories across the evaluation benchmarks.}
\vspace{-4mm}
\label{fig:trajectory_statistics}
\end{figure*}

\textbf{Real-world print-and-recapture evaluation.}
We construct a real-world evaluation set by printing the optimized texture at a fixed physical size and attaching it to an approximately planar surface.
For each sample, we record a clean reference video, an adversarial video with the optimized texture, and a calibration video with a uniform white patch of the same size and attachment location.
The white-patch recordings provide a high-contrast reference for estimating frame-wise homographies and visibility masks. 
These annotations enable future studies to evaluate adversarial textures under identical real-world observation trajectories, ensuring reproducible comparisons. %
We further conduct an independent human evaluation to confirm that the underlying actions remain semantically interpretable. 
Together, the annotated benchmark videos and real-world captures establish a reusable and reproducible testbed for evaluating physical robustness under dynamic observation conditions.

\textbf{Dataset statistics and reusable benchmark.}
Table~\ref{tab:trajectory_stats} summarizes the overall characteristics of the constructed benchmark, comprising 605 clips (19,251 frames) from EK100, 328 clips (10,204 frames) from SSV2, and 348 real-world clips (8,645 frames).
In Table~\ref{tab:trajectory_stats}, homography coverage measures the fraction of frames with valid surface projections; mask rate measures the frequency of occlusion within the projected attachment region; texture area denotes the average fraction of the image occupied by the projected texture; and motion range quantifies the spatial extent of its trajectory.
\figureautorefname~\ref{fig:trajectory_statistics} further characterizes the benchmark at the frame level, illustrating the spatial distribution of projected texture centers (a), variations in texture geometry and visibility (b), and the distributions of visible texture fraction (c) and projected texture area (d).
Together, these statistics characterize the diverse observation conditions captured by our benchmark.
For reproducible comparison of different attack methods, by fixing $\gamma_i$, a new texture $\bm{\delta}'$ can be rendered to construct $\tilde{\bm{X}}_i(\bm{\delta}';\gamma_i)$ under identical geometry and visibility.
More details such as full benchmark construction and validation pipeline, the quality-control criteria, and metric definitions are provided in the Appendix~\ref{app:surface_trajectory}.

\vspace{-1mm}
\section{Experimental Results}
\vspace{-2mm}

\begin{table}[t]
\centering
\vspace{-1mm}
\caption{Comparison performance on EK100 future action anticipation. Top1 ASR (\%, $\uparrow$) is reported on our split Train and Val data, with best and second-best results highlighted in bold and underlined.}
\vspace{-1mm}
\label{tab:main-ek100}
\renewcommand{\arraystretch}{1}
\resizebox{0.87\columnwidth}{!}{%
\begin{tabular}{c|cc|cc|cc|cc}
\toprule
Models & \multicolumn{2}{c|}{V-JEPA2-L} & \multicolumn{2}{c|}{V-JEPA2.1-B} & \multicolumn{2}{c|}{AVT-B} & \multicolumn{2}{c}{Overall} \\
\cmidrule(r){1-1} \cmidrule(r){2-3} \cmidrule(r){4-5} \cmidrule(r){6-7} \cmidrule(r){8-9}
Setting & Train & Val & Train & Val & Train & Val & Avg. Train & Avg. Val \\
\midrule
Black texture  & 4.8 & 3.3 & 5.7 & 5.0 & 8.2 & 5.0 & 6.2 & 4.4 \\
Random texture & 5.5 & 4.7 & 8.9 & 5.5 & 7.4 & 6.1 & 7.3 & 5.4 \\
\midrule
EOT        & \underline{30.3} & \underline{25.1} & \underline{59.3} & \underline{57.1} & 32.9 & 25.0 & \underline{40.8} & \underline{35.7} \\
Patch-Fool & 6.5 & 9.0 & 46.9 & 27.9 & 28.2 & 18.3 & 27.2 & 18.4 \\
UPA-RFAS   & 9.5 & 10.4 & 17.1 & 15.7 & 17.6 & 18.3 & 14.7 & 14.8 \\
BTC-UAP    & {13.9} & 10.4 & 11.4 & 13.6 & 36.5 & \underline{28.3} & 20.6 & 17.4 \\
PhyCamo    & {13.9} & 13.7 & 58.3 & 29.3 & \textbf{42.4} & \textbf{30.0} & 38.2 & 24.3 \\
\midrule
\textbf{TARA} & \textbf{37.7} & \textbf{26.1} & \textbf{70.6} & \textbf{67.1} & \underline{37.6} & \textbf{30.0} & \textbf{48.6} & \textbf{41.1} \\
\bottomrule
\end{tabular}
}
\vspace{-1mm}
\end{table}

\begin{table}[t]
\centering
\caption{Comparison performance on SSV2 video action recognition. Top1 ASR (\%, $\uparrow$) is reported on our split Train and Val data, with best and second-best results highlighted in bold and underlined.}
\vspace{-1mm}
\label{tab:main-ssv2-recognition}
\renewcommand{\arraystretch}{1}
\resizebox{0.87\columnwidth}{!}{%
\begin{tabular}{c|cc|cc|cc|cc}
\toprule
Models & \multicolumn{2}{c|}{V-JEPA2-L} & \multicolumn{2}{c|}{VideoMAE-B} & \multicolumn{2}{c|}{InternVideo2} & \multicolumn{2}{c}{Overall} \\
\cmidrule(r){1-1} \cmidrule(r){2-3} \cmidrule(r){4-5} \cmidrule(r){6-7} \cmidrule(r){8-9}
Setting & Train & Val & Train & Val & Train & Val & Avg. Train & Avg. Val \\
\midrule
Black texture  & 9.2 & 5.6 & 5.7 & 10.8 & 4.0 & 2.9 & 6.3 & 6.4 \\
Random texture & 7.4 & 4.9 & 1.1 & 1.8 & 2.1 & 0.3 & 3.5 & 2.3 \\
\midrule
EOT        & \underline{25.5} & \underline{30.6} & 14.9 & \underline{16.9} & 15.8 & 13.2 & {18.7} & \underline{20.2} \\
Patch-Fool & 12.5 & 8.3 & 13.7 & \underline{16.9} & 5.1 & 2.2 & 10.4 & 9.1 \\
UPA-RFAS   & 9.8 & 3.5 & 14.3 & \underline{16.9} & 11.9 & 9.6 & 12.0 & 10.0 \\
BTC-UAP    & 12.5 & 5.6 & \underline{15.4} & \textbf{17.7} & 14.1 & 15.7 & 14.0 & 13.0 \\
PhyCamo    & 17.9 & 12.5 & \underline{15.4} & \underline{16.9} & \underline{24.3} & \underline{23.5} &\underline{19.2} & 17.6 \\
\midrule
\textbf{TARA} & \textbf{38.6} & \textbf{34.7} & \textbf{17.1} & \textbf{17.7} & \textbf{41.2} & \textbf{25.7} & \textbf{32.3} & \textbf{26.0} \\
\bottomrule
\end{tabular}
}
\vspace{-5mm}
\end{table}

\subsection{Experimental Setup}
\vspace{-2mm}
\textbf{Evaluation benchmarks and models.}
We evaluate our attack across three tasks using the geometry-grounded EK100~\citep{damen2022rescaling}, SSV2~\citep{goyal2017something} and our real-world captured benchmarks.
The evaluated tasks include:
\begin{itemize}
\vspace{-2mm}
    \item \textbf{Future action anticipation} on EK100 using V-JEPA2-L~\citep{assran2025v}, V-JEPA2.1-B~\citep{mur2026v}, and AVT-B~\citep{girdhar2021anticipative}, and our real-world captured action anticipation set, using V-JEPA2-L and V-JEPA2.1-B. 
    \vspace{-1mm}
    \item \textbf{Action recognition} on SSV2, using V-JEPA2-L, VideoMAE-B~\citep{tong2022videomae}, and InternVideo2-dist-L~\citep{wang2024internvideo2}, and our real-world captured action recognition set, using V-JEPA2-L and InternVideo2-dist-L. 
    \vspace{-1mm}
    \item \textbf{Action-conditioned future frame generation} on SSV2 using DiLA~\citep{zhang2026dila}.
    \vspace{-4mm}
\end{itemize}
\textbf{Evaluation metrics and data splits.}
For action anticipation and recognition tasks, we report Top-1 attack success rate (ASR), defined as the percentage of initially correct predictions that become incorrect under attack.
For future frame generation, we report PSNR and MSE against the clean target frames, where lower PSNR and higher MSE relative to clean reconstruction indicate greater attack-induced degradation.
We partition the 605 EK100 clips into 356 training and 249 validation clips, and the 328 SSV2 clips into 184 and 144, respectively.
The adversarial texture is optimized utilizing only the training videos and their associated surface trajectories.
The held-out validation clips provide unseen samples for assessing the generalization of adversarial texture across different videos and object--camera trajectories.

\textbf{Compared methods and implementation details.}
We compare TARA with representative adversarial patch and physical attack methods.
Specifically, we consider adversarial patch-based methods, including patch with expectation over transformation (EOT)~\citep{brown2017adversarial}, the attention-guided Patch-Fool~\citep{fu2022patchfool}, and PhyCamo~\citep{zhang2025phycamo}, which employs contrastive learning to improve the robustness of physical camouflage across viewpoints.
We further include BTC-UAP~\citep{kim2023breaking}, which focuses on temporal feature disruption, and UPA-RFAS~\citep{lu2026robots}, which employs a visual feature-space objective to optimize transferable physical patches for vision-language-action models.
For a fair comparison, all compared methods incorporate trajectory-aware optimization using the same annotated object--camera trajectories and geometric constraints described in Section~\ref{sec:pre}.
Black and random textures are included as non-optimized controls to distinguish adversarial optimization from the effects of ordinary texture insertion.
For each victim model, all methods optimize a universal circular texture with a radius of 24, using the same training and evaluation splits.
We preserve each model's native temporal sampling, spatial preprocessing, and prediction aggregation.
The details of the model- and task-specific adaptations, optimization schedules and budgets, checkpoint-selection criteria, random seeds, and hardware configurations are provided in the Appendix~\ref{app:implementation_details}.

\vspace{-3mm}
\subsection{Main Results on geometry-grounded EK100 and SSV2}
\vspace{-2mm}
\textbf{Performance on EK100 action anticipation.}
Table~\ref{tab:main-ek100} compares the attack performance of different methods on EK100 future action anticipation across V-JEPA2-L, V-JEPA2.1-B, and AVT-B, with Top-1 ASR reported.
In particular, TARA achieves the highest average ASR of 48.6\% and 41.1\% on the training and validation splits, outperforming the strongest baseline by 7.8\% and 5.4\%, respectively.
Notably, V-JEPA2.1-B exhibits substantially higher adversarial vulnerability than V-JEPA2-L across several attack methods. 
Although V-JEPA2.1 incorporates dense feature modeling to capture finer-grained spatiotemporal information, its representations may also be more sensitive to persistent, localized adversarial perturbations.
These findings suggest that richer spatiotemporal representations do not necessarily translate into greater robustness against persistent physical attacks.
Overall, these results demonstrate TARA's effectiveness in disrupting future action anticipation across distinct video architectures and its generalization to unseen video instances under dynamic observation conditions.

\textbf{Performance on SSV2 action recognition.}
Table~\ref{tab:main-ssv2-recognition} compares the attack performance of different methods on SSV2 action recognition across V-JEPA2-L, VideoMAE-B, and InternVideo2, with Top-1 ASR reported.
It indicates that TARA achieves the highest average ASR of 32.3\% and 26.0\% on the training and validation splits, outperforming the strongest baseline by 13.1\% and 5.8\%, respectively.
The improvements are particularly pronounced on InternVideo2, where TARA achieves 41.2\% and 25.7\% ASR, surpassing PhyCamo by 16.9\% and 2.2\%, respectively.
In contrast, the performance gains on VideoMAE-B are relatively modest, with TARA achieving 17.1\% and 17.7\% ASR, suggesting that different video representation architectures exhibit varying susceptibility to persistent physical perturbations.
Overall, TARA maintains the highest average ASR across both splits, indicating that its trajectory-aware optimization can effectively accommodate heterogeneous video representations while preserving attack effectiveness under dynamic observation conditions.

\begin{wraptable}{r}{0.5\columnwidth}
\centering
\vspace{-2mm}
\caption{Attack performance on SSV2 future frame generation using DiLA.}
\label{tab:main-ssv2-reconstruction}
\vspace{-2mm}
\resizebox{0.5\columnwidth}{!}{%
\begin{tabular}{l|cc|cc}
\toprule
& \multicolumn{2}{c|}{Train}
& \multicolumn{2}{c}{Val} \\
\cmidrule(r){2-3} \cmidrule(l){4-5}
Method & PSNR$\downarrow$ & MSE$(\times 10^{-3})\uparrow$
       & PSNR$\downarrow$ & MSE$(\times 10^{-3})\uparrow$ \\
\midrule
Clean & 17.37 & 24.42 & 17.19 & 24.33 \\
Black patch& 16.48 & 27.70 & 16.51 & 26.74 \\
Random patch& 17.10 & 24.88 & 17.12 & 24.37 \\
\midrule
EOT & 16.30 & 29.96 & 16.25 & 29.03 \\
Patch-Fool & 15.94 & 33.83 & 16.41 & 28.48 \\
UPA-RFAS & \underline{15.32} & \underline{34.47}
         & \underline{15.58} & \underline{33.27} \\
BTC-UAP & 16.83 & 25.07 & 16.78 & 25.54 \\
\midrule
\textbf{TARA}
& \textbf{11.38} & \textbf{90.38}
& \textbf{12.66} & \textbf{67.37} \\
\bottomrule
\end{tabular}%
}
\vspace{-3mm}
\end{wraptable}
\textbf{Performance on SSV2 future frame generation.} Table~\ref{tab:main-ssv2-reconstruction} compares the attack performance of different methods on latent-action-conditioned future frame generation using DiLA, with reconstruction PSNR and MSE reported on both the training and validation splits.
It indicates that TARA achieves the lowest PSNR of 11.38 and 12.66 dB and the highest MSE of 90.38 and 67.37 ($\times 10^{-3}$) on the training and validation splits, respectively, indicating substantial degradation in future frame reconstruction quality.
Compared with the strongest baseline, UPA-RFAS, TARA further reduces PSNR by 3.94 and 2.92 dB while increasing MSE by 55.91 and 34.10 ($\times 10^{-3}$), respectively.
These results indicate that TARA can substantially disrupt latent-action-conditioned future frame generation, extending its effectiveness beyond discriminative action recognition and anticipation to generative video prediction.
Attacking objective and visualization for the generation task are provided in the Appendix~\ref{app: attack for dila}.

\begin{table}[t]
\centering
\vspace{-2mm}
\caption{Performance on real-world captured data. The first six rows are digital projection results using recorded homographies and masks; the last two rows are physical print-and-recapture results.}
\label{tab:real-world}
\renewcommand{\arraystretch}{1}
\resizebox{0.8\columnwidth}{!}{%
\begin{tabular}{c|cc|cc}
\toprule
Tasks & \multicolumn{2}{c|}{Action Anticipation} & \multicolumn{2}{c}{Action Recognition} \\
\cmidrule(r){1-1} \cmidrule(r){2-3} \cmidrule(r){4-5}
Models & V-JEPA2-L & V-JEPA2.1-B & V-JEPA2-L & InternVideo2 \\
\midrule
EOT        & \underline{29.7} & \textbf{58.8} & \underline{21.2} & 40.3 \\
Patch-Fool & 17.6 & 23.5 & 10.6 & 35.5 \\
UPA-RFAS   & 18.9 & 17.6 & 11.8 & \underline{43.5} \\
BTC-UAP    & 14.9 & 17.6 & 12.9 & 24.2 \\
PhyCamo    & 13.5 & \underline{29.4} & 17.6 & 38.7 \\
\midrule
\textbf{TARA} & \textbf{32.4} & \textbf{58.8} & \textbf{25.8} & \textbf{50.0} \\
\midrule
\textit{Physical EOT (printed)} & 28.6 & \textbackslash & 23.1 & \textbackslash \\
\textit{Physical TARA (printed)} & 32.1 & \textbackslash & 28.2 & \textbackslash \\
\bottomrule
\end{tabular}
}
\end{table}
\begin{figure*}[t]
\vspace{-3mm}
\centering
\includegraphics[width=0.75\textwidth]{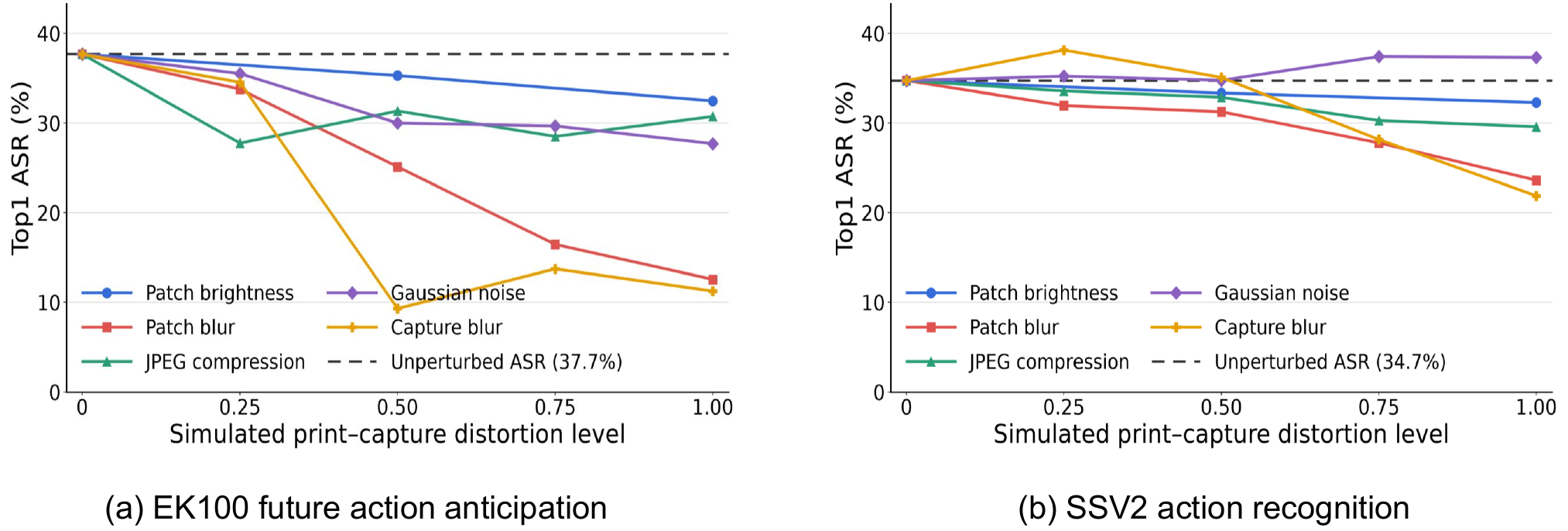}
\vspace{-1mm}
\caption{Performance of TARA under varying levels of simulated print-and-capture distortions.}
\vspace{-3mm}
\label{fig:performance-distortion-curve_v1}
\vspace{-3mm}
\end{figure*}

\vspace{-2mm}
\subsection{Real-world Results and distortion analysis}
\vspace{-2mm}
\textbf{Performance on real-world captured data.}
Table~\ref{tab:real-world} compares the attack performance of different methods on our real-world captured data.
We first evaluate existing methods by digitally projecting their adversarial textures onto video-trajectory instances using our recorded homographies and masks.
It indicates that TARA achieves the highest or joint-highest ASR across all evaluated models and tasks.
In physical print-and-recapture experiments, TARA achieves 32.1\% and 28.2\% ASR on V-JEPA2-L for action anticipation and recognition, respectively, outperforming EOT by 3.5\% and 5.1\%.
These results demonstrate that TARA maintains its attack effectiveness under uncontrolled object--camera trajectories and additional print-and-capture distortions, supporting its effectiveness under real-world physical deployment.

\textbf{Robustness to simulated print--capture distortions.}
To further examine its sensitivity to print-and-capture distortions, \figureautorefname~\ref{fig:performance-distortion-curve_v1} reports the attack performance of our TARA texture under varying levels of distortions, including brightness changes, Gaussian noise, blur, and JPEG compression.
TARA maintains relatively stable ASR under brightness variations, Gaussian noise, and JPEG compression, while stronger patch and capture blur lead to performance degradation, particularly on EK100 action anticipation.
This degradation may result from the smoothing of local adversarial texture structures and the loss of fine-grained visual details, as further illustrated by the distorted texture visualizations in the Appendix~\ref{app:more visual}.
Overall, these results complement our physical print-and-recapture evaluation, demonstrating that TARA maintains its attack effectiveness under a wide range of realistic imaging distortions within practical operating conditions.

\vspace{-2mm}
\subsection{Ablation Study and Analysis}
\vspace{-2mm}
\begin{table}[t]
\centering
\caption{Ablation study of our TARA on the EK100 dataset. Top-1 ASR (\%, $\uparrow$) is reported. TA, CA, and TR denote trajectory awareness, contrastive alignment, and task refinement, respectively.}
\vspace{-1mm}
\label{tab:ablation-ek100}
\resizebox{0.75\columnwidth}{!}{%
\begin{tabular}{ccc|cc|cc|cc|cc}
\toprule
\multicolumn{3}{c|}{Components} & \multicolumn{2}{c|}{V-JEPA2-L} & \multicolumn{2}{c|}{V-JEPA2.1-B} & \multicolumn{2}{c|}{AVT-B} & \multicolumn{2}{c}{Overall} \\
\cmidrule(lr){1-3} \cmidrule(lr){4-5} \cmidrule(lr){6-7} \cmidrule(lr){8-9} \cmidrule(l){10-11}
TA & CA & TR & Train & Val & Train & Val & Train & Val & Avg. Train & Avg. Val \\
\midrule
\xmark & \cmark & \cmark & 10.4 & 10.9 & 8.0 & 5.7 & 9.4 & 3.3 & 9.3 & 6.6 \\
\cmark & \xmark & \cmark & 32.9 & 22.7 & 68.6 & 58.3 & \textbf{40.0} & 28.3 &  47.2 & 36.4 \\
\cmark & \cmark & \xmark & 29.0 & 22.7 & 64.0 & 65.0 & 36.5 & \textbf{30.0} & 43.2 & 39.2 \\
\midrule
\cmark & \cmark & \cmark & \textbf{37.7} & \textbf{26.1} & \textbf{70.6} & \textbf{67.1} & 37.6 & \textbf{30.0} & \textbf{48.6} & \textbf{41.1} \\
\bottomrule
\end{tabular}%
}
\vspace{-1mm}
\end{table}
\textbf{Ablation study.}
Table~\ref{tab:ablation-ek100} investigates the contributions of each component.
To verify the contribution of trajectory awareness learning, we optimize a universal adversarial patch at a fixed image-center location without using the annotated video trajectories.
The results highlight trajectory-aware optimization as a key component of TARA, contributing substantial improvements in attack performance.
Meanwhile, both contrastive alignment and task refinement contribute non-trivial performance gains, further improving TARA's overall attack effectiveness.
Together, these findings confirm the contributions of all three components to TARA's overall performance.

\begin{figure}[t]
\vspace{-1mm}
    \centering
    \includegraphics[width=0.9\linewidth]{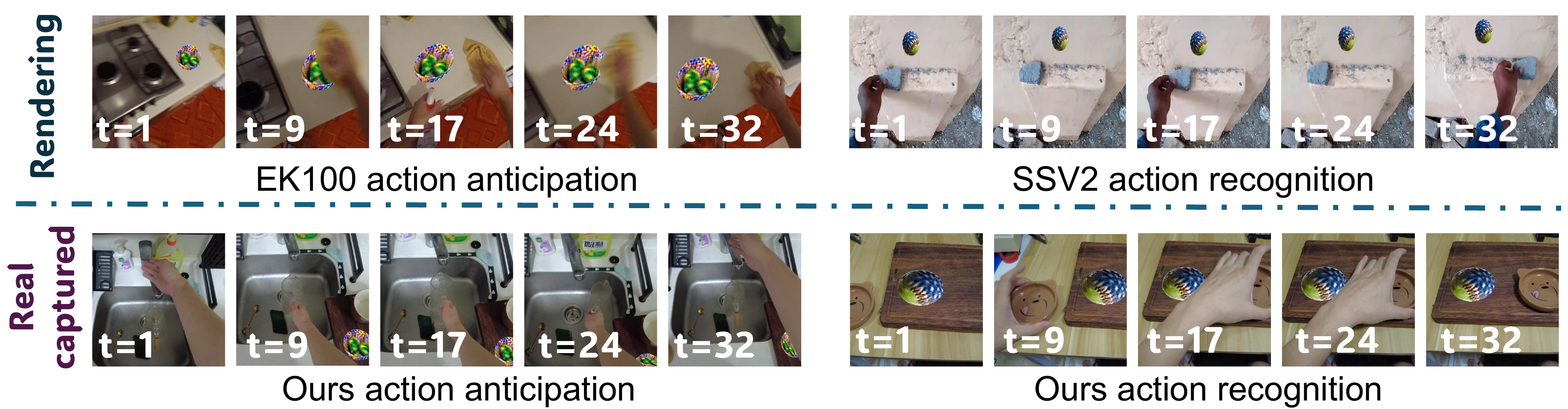}
    \vspace{-2mm}
    \caption{Visualization of TARA texture. The top row shows geometry-grounded texture rendering on EK100 and SSV2 videos, and the bottom row presents physical print-and-recapture examples.}
    \label{fig:visual}
    \vspace{-6mm}
\end{figure}

\textbf{Visualization of adversarial textures.}
\figureautorefname~\ref{fig:visual} presents representative video frames of TARA under geometry-grounded rendering and physical print-and-recapture settings for both action anticipation and recognition tasks.
The adversarial texture undergoes substantial variations in viewpoint, scale, position, and visibility along diverse object--camera trajectories, illustrating the dynamic observation conditions considered in our evaluation.
The real-world captures further demonstrate how a fixed printed texture appears across different physical scenarios under uncontrolled observation conditions.
Additional visualizations covering diverse video instances and observation trajectories are provided in the Appendix~\ref{app:more visual} and~\ref{sec:full traj}.

\textbf{Generalization across content and trajectories.}
We further evaluate TARA by independently varying video content and observation trajectories, including unseen trajectories from independently captured real-world videos.
We find that TARA retains its attack effectiveness under unseen trajectories, even when applied to unseen video content, demonstrating its generalization beyond the training video--trajectory instances. 
The detailed results are in Appendix~\ref{app:content-trajectory-generalization}.

\textbf{Statistical Stability.} We further evaluate the statistical stability of TARA across three random seeds, observing stable performance on both EK100 and SSV2. Details are shown in Appendix~\ref{app:seed_stability}.

\vspace{-2mm}
\section{Related Work}
\vspace{-2mm}
\textbf{Video understanding models and world models.}
Video pretraining learns transferable spatiotemporal representations for understanding scenes and actions in videos.
Early research utilizes reconstruction-based objectives, such as VideoMAE and VideoMAE V2~\citep{tong2022videomae,wang2023videomae}.
Then, InternVideo2 and VideoPrism~\citep{wang2024internvideo2,zhao2024videoprism} scale training data and supervision toward more general video understanding.
Beyond interpreting observed events, AVT predicts future action labels from preceding observations~\citep{girdhar2021anticipative}, while world models such as iVideoGPT and DINO-WM~\citep{wu2024ivideogpt,zhou2025dino} predict action-conditioned future observations or visual features to support planning.
V-JEPA instead learns representations by predicting masked latent video features~\citep{bardes2024revisiting} and V-JEPA 2~\citep{assran2025v} connects these capabilities for video understanding and anticipation.
Their reliance on visual observations motivates investigating their physical robustness to camera input.

\textbf{Digital adversarial attacks on video and world models.}
%
Early video attacks exploit temporal sparsity and image-to-video feature transfer~\citep{wei2019sparse,wei2022cross}, while recent work targets pretrained video representations and transfers perturbations across downstream tasks~\citep{xia2024transferable,lu2026pretrain}.
World-model attacks further explore structured conditioning perturbations, automated attack search against closed-loop agents, and poisoning through generated data~\citep{guo2026world,guo2026wmattack,rathbun2026targeting}.
These approaches operate through digital access to inputs, conditioning signals, or training pipelines.
Our setting instead constrains the adversarial input to a single printable texture in real-world scenarios, with its observation governed by object--camera trajectories.
The texture must therefore generalize jointly across video semantics and evolving observations.

\textbf{Physical adversarial attacks.}
Early work established printable adversarial patches and transformation-based optimization against fabrication and imaging variations~\citep{brown2017adversarial,athalye2018synthesizing,eykholt2018robust}.
Subsequent studies extended physical attacks to videos and tracking by modeling temporal and motion-induced variations~\citep{pony2021over,du2022motional,long2024papmot}, and improved robustness across poses and viewpoints through multi-view or geometry-aware optimization~\citep{li2025uv,zhang2025phycamo,lou20253d}.
Recent work further attacks embodied policies, including diffusion policies and VLA models, with efforts toward cross-model transferability and partial trajectory observability~\citep{chen2024diffusion,wang2025exploring,lu2026robots,wang2026partially}.
Tex3D and UniTexture optimize object-bound textures along manipulation trajectories and across multiple VLA tasks~\citep{chen2026tex3d,dai2026unitexture}.
These studies improve temporal persistence, multi-view robustness, transferability, or geometric consistency, but do not study joint generalization across heterogeneous video semantics and unseen object--camera trajectories.

\vspace{-3mm}
\section{Conclusion}
\vspace{-3mm}
In this work, we investigate a practical physical attack setting: a single printable adversarial texture across heterogeneous videos and uncontrolled object--camera trajectories.
We identify conflicting optimization gradients across diverse video--trajectory instances as a key challenge to generating an effective adversarial texture.
To address this challenge, we propose TARA, which employs trajectory-aware contrastive alignment to encourage shared adversarial representations across instances, followed by trajectory-aware task refinement to induce task-level prediction errors.
We further establish a reproducible geometry-grounded benchmark with temporally consistent homography and visibility annotations, and conduct real-world print-and-recapture experiments to validate attack effectiveness under practical deployment conditions.
Extensive evaluations in both simulated and real-world print-and-recapture experiments demonstrate that TARA effectively disrupts video understanding and anticipation across unseen video--trajectory instances, highlighting the security risks posed by persistent physical attacks in real-world deployments.


\subsection*{AI use statement}
In this work, we did not use generative AI tools for tasks requiring mandatory disclosure. We have not used generative AI tools to develop theoretical models, formulate mathematical claims, design experiments, or interpret results, and the remaining mandatory-disclosure tasks are not applicable to this work. Additionally, we used generative AI tools to create or edit software code and to proofread the manuscript for grammar, clarity, and readability. We have reviewed all AI-assisted work. The authors reviewed and tested AI-assisted code for correctness and verified that the suggested revisions preserved the technical meaning and accuracy of the manuscript. We take responsibility for the final content of this work, including text, claims, or artifacts produced with the aid of generative AI.

\subsection*{Ethics Statement}
This work investigates the vulnerability of world models and video understanding models to persistent physical adversarial textures. Our objective is to characterize this security risk and support the development of more robust models. We evaluate the attack on established video benchmarks and through controlled print-and-recapture experiments, rather than deploying it in operational systems. Our findings highlight the need to assess model robustness under temporally varying physical observations, particularly before deploying video-based models in safety-critical applications.

\subsection*{Reproducibility Statement}
We provide detailed descriptions of the attack formulation, optimization procedure, and evaluation protocols in the main paper, with additional implementation settings, trajectory-construction procedures, and quality-control details in the Appendix.
To facilitate reproducibility, we will publicly release the code and data required to reproduce all experiments in this work, including our generated trajectory annotations and real-world captured videos.
These resources will support the reproduction of our results and enable future research to evaluate adversarial textures under various video--trajectory conditions.

\bibliography{iclr2027_conference}

@String(IJCV = {Int. J. Comput. Vis.})

@String(CVPR= {IEEE Conf. Comput. Vis. Pattern Recog.})

@String(ICCV= {Int. Conf. Comput. Vis.})

@String(ECCV= {Eur. Conf. Comput. Vis.})

@String(NIPS= {Adv. Neural Inform. Process. Syst.})

@String(ICLR = {Int. Conf. Learn. Represent.})

@String(AAAI = {AAAI})

@String(ICML = {Int. Conf. Machine Learning})

@article{brown2017adversarial,
  title={Adversarial patch},
  author={Brown, Tom B and Man{\'e}, Dandelion and Roy, Aurko and Abadi, Mart{\'\i}n and Gilmer, Justin},
  journal={arXiv preprint arXiv:1712.09665},
  year={2017}
}

@inproceedings{athalye2018synthesizing,
  title={Synthesizing robust adversarial examples},
  author={Athalye, Anish and Engstrom, Logan and Ilyas, Andrew and Kwok, Kevin},
  booktitle=icml,
  year={2018},
}

@inproceedings{eykholt2018robust,
  title={Robust physical-world attacks on deep learning visual classification},
  author={Eykholt, Kevin and Evtimov, Ivan and Fernandes, Earlence and Li, Bo and Rahmati, Amir and Xiao, Chaowei and Prakash, Atul and Kohno, Tadayoshi and Song, Dawn},
  booktitle=cvpr,
  year={2018},
}

@article{du2022motional,
  title={A motional but temporally consistent physical video examples},
  author={Du, Zhenyu and Wei, Xingxing and Zhang, Weiming and Liu, Fangzheng and Bian, Huanyu and Liu, Jiayang},
  journal={Journal of Information Security and Applications},
  year={2022},
}

@inproceedings{pony2021over,
  title={Over-the-air adversarial flickering attacks against video recognition networks},
  author={Pony, Roi and Naeh, Itay and Mannor, Shie},
  booktitle=cvpr,
  year={2021}
}

@article{assran2025v,
  title={V-jepa 2: Self-supervised video models enable understanding, prediction and planning},
  author={Assran, Mido and Bardes, Adrien and Fan, David and Garrido, Quentin and Howes, Russell and Muckley, Matthew and Rizvi, Ammar and Roberts, Claire and Sinha, Koustuv and Zholus, Artem and others},
  journal={arXiv preprint arXiv:2506.09985},
  year={2025}
}

@article{tong2022videomae,
  title={Videomae: Masked autoencoders are data-efficient learners for self-supervised video pre-training},
  author={Tong, Zhan and Song, Yibing and Wang, Jue and Wang, Limin},
  journal=nips,
  year={2022}
}

@inproceedings{wang2023videomae,
  title={Videomae v2: Scaling video masked autoencoders with dual masking},
  author={Wang, Limin and Huang, Bingkun and Zhao, Zhiyu and Tong, Zhan and He, Yinan and Wang, Yi and Wang, Yali and Qiao, Yu},
  booktitle=CVPR,
  year={2023},
}

@article{zhao2024videoprism,
  title={Videoprism: A foundational visual encoder for video understanding},
  author={Zhao, Long and Gundavarapu, Nitesh B and Yuan, Liangzhe and Zhou, Hao and Yan, Shen and Sun, Jennifer J and Friedman, Luke and Qian, Rui and Weyand, Tobias and Zhao, Yue and others},
  journal=icml,
  year={2024}
}

@inproceedings{wang2024internvideo2,
  title={Internvideo2: Scaling foundation models for multimodal video understanding},
  author={Wang, Yi and Li, Kunchang and Li, Xinhao and Yu, Jiashuo and He, Yinan and Chen, Guo and Pei, Baoqi and Zheng, Rongkun and Wang, Zun and Shi, Yansong and others},
  booktitle=eccv,
  year={2024},
}

@article{zhou2025dino,
  title={Dino-wm: World models on pre-trained visual features enable zero-shot planning},
  author={Zhou, Gaoyue and Pan, Hengkai and LeCun, Yann and Pinto, Lerrel},
  journal=ICML,
  year={2025}
}

@article{wu2024ivideogpt,
  title={ivideogpt: Interactive videogpts are scalable world models},
  author={Wu, Jialong and Yin, Shaofeng and Feng, Ningya and He, Xu and Li, Dong and Hao, Jianye and Long, Mingsheng},
  journal=nips,
  year={2024}
}

@article{xia2024transferable,
  title={Transferable adversarial attacks on sam and its downstream models},
  author={Xia, Song and Yang, Wenhan and Yu, Yi and Lin, Xun and Ding, Henghui and Duan, Lingyu and Jiang, Xudong},
  journal=nips,
  year={2024}
}

@article{ding2026transferable,
  title={Transferable Adversarial Attack on Referring Video Object Segmentation},
  author={Ding, Meiwen and Xia, Song and Yu, Yi and He, Shuting and Jiang, Xudong},
  journal={IEEE Transactions on Information Forensics and Security},
  year={2026},
}

@inproceedings{wei2019sparse,
  title={Sparse adversarial perturbations for videos},
  author={Wei, Xingxing and Zhu, Jun and Yuan, Sha and Su, Hang},
  booktitle=AAAI,
  year={2019}
}

@inproceedings{wei2022cross,
  title={Cross-modal transferable adversarial attacks from images to videos},
  author={Wei, Zhipeng and Chen, Jingjing and Wu, Zuxuan and Jiang, Yu-Gang},
  booktitle=cvpr,
  year={2022},
}

@inproceedings{lu2026pretrain,
  title={From pretrain to pain: Adversarial vulnerability of video foundation models without task knowledge},
  author={Lu, Hui and Yu, Yi and Xia, Song and Yang, Yiming and Rajan, Deepu and Ng, Boon Poh and Kot, Alex and Jiang, Xudong},
  booktitle=AAAI,
  year={2026}
}

@inproceedings{long2024papmot,
  title={Papmot: Exploring adversarial patch attack against multiple object tracking},
  author={Long, Jiahuan and Jiang, Tingsong and Yao, Wen and Jia, Shuai and Zhang, Weijia and Zhou, Weien and Ma, Chao and Chen, Xiaoqian},
  booktitle=eccv,
  year={2024},
}

@article{chen2024diffusion,
  title={Diffusion policy attacker: Crafting adversarial attacks for diffusion-based policies},
  author={Chen, Yipu and Xue, Haotian and Chen, Yongxin},
  journal=nips,
  year={2024}
}

@inproceedings{li2025uv,
  title={UV-attack: Physical-world adversarial attacks on person detection via dynamic-NeRF-based UV mapping},
  author={Li, Yanjie and Liang, Kaisheng and Xiao, Bin},
  booktitle=ICLR,
  year={2025}
}

@inproceedings{wang2025exploring,
  title={Exploring the adversarial vulnerabilities of vision-language-action models in robotics},
  author={Wang, Taowen and Han, Cheng and Liang, James and Yang, Wenhao and Liu, Dongfang and Zhang, Luna Xinyu and Wang, Qifan and Luo, Jiebo and Tang, Ruixiang},
  booktitle=iccv,
  year={2025},
}

@inproceedings{hingun2023reap,
  title={REAP: A large-scale realistic adversarial patch benchmark},
  author={Hingun, Nabeel and Sitawarin, Chawin and Li, Jerry and Wagner, David},
  booktitle=iccv,
  year={2023}
}

@inproceedings{braunegg2020apricot,
  title={Apricot: A dataset of physical adversarial attacks on object detection},
  author={Braunegg, Anneliese and Chakraborty, Amartya and Krumdick, Michael and Lape, Nicole and Leary, Sara and Manville, Keith and Merkhofer, Elizabeth and Strickhart, Laura and Walmer, Matthew},
  booktitle=eccv,
  pages={35--50},
  year={2020},
  organization={Springer}
}

@inproceedings{goyal2017something,
  title={The “something something” video database for learning and evaluating visual common sense},
  author={Goyal, Raghav and Kahou, Samira Ebrahimi and Michalski, Vincent and Materzynska, Joanna and Westphal, Susanne and Kim, Heuna and Haenel, Valentin and Fruend, Ingo and Yianilos, Peter and Mueller-Freitag, Moritz and others},
  booktitle=iccv,
  year={2017}
}

@article{damen2022rescaling,
  title={Rescaling egocentric vision: Collection, pipeline and challenges for epic-kitchens-100},
  author={Damen, Dima and Doughty, Hazel and Farinella, Giovanni Maria and Furnari, Antonino and Kazakos, Evangelos and Ma, Jian and Moltisanti, Davide and Munro, Jonathan and Perrett, Toby and Price, Will and others},
  journal=ijcv,
  pages={33--55},
  year={2022},
}

@article{
bardes2024revisiting,
title={Revisiting Feature Prediction for Learning Visual Representations from Video},
author={Adrien Bardes and Quentin Garrido and Jean Ponce and Xinlei Chen and Michael Rabbat and Yann LeCun and Mido Assran and Nicolas Ballas},
journal={Transactions on Machine Learning Research},
year={2024}
}

@inproceedings{girdhar2021anticipative,
  title={Anticipative video transformer},
  author={Girdhar, Rohit and Grauman, Kristen},
  booktitle=iccv,
  year={2021}
}

@article{guo2026world,
  title={When world models dream wrong: Physical-conditioned adversarial attacks against world models},
  author={Guo, Zhixiang and Liang, Siyuan and Balogh, Andras and Lunberry, Noah and Tu, Rong-Cheng and Jelasity, Mark and Tao, Dacheng},
  journal={arXiv preprint arXiv:2602.18739},
  year={2026}
}

@article{guo2026wmattack,
  title={WMAttack: Automated Attack Search for Adversarial Evaluation of World-Model Agents},
  author={Guo, Zhixiang and Liang, Siyuan and Fu, Shi and Guo, Cheng and Balogh, Andras and Jelasity, Mark and Tao, Dacheng},
  journal={arXiv preprint arXiv:2605.23220},
  year={2026}
}

@article{rathbun2026targeting,
  title={Targeting World Models to Compromise Robot Learning Pipelines},
  author={Rathbun, Ethan and Agha, Ahmed and Mahmud, Saaduddin and Amato, Christopher and Oprea, Alina and Bagdasarian, Eugene},
  journal={arXiv preprint arXiv:2606.09499},
  year={2026}
}

@article{chen2026tex3d,
  title={Tex3D: Objects as attack surfaces via adversarial 3D textures for vision-language-action models},
  author={Chen, Jiawei and Huang, Simin and Du, Jiawei and Chen, Shuaihang and Tian, Yu and Wei, Mingjie and Yu, Chao and Yin, Zhaoxia},
  journal={arXiv preprint arXiv:2604.01618},
  year={2026}
}

@article{dai2026unitexture,
  title={UniTexture: Cross-Task Universal Adversarial Textures for Vision-Language-Action Models},
  author={Dai, Yukun and Dai, Mingzhe and Wang, Tianshi and Li, Fengling and Li, Jingjing and Zhu, Lei},
  journal={arXiv preprint arXiv:2608.13453},
  year={2026}
}

@inproceedings{lu2026robots,
  title={When robots obey the patch: Universal transferable patch attacks on vision-language-action models},
  author={Lu, Hui and Yu, Yi and Yang, Yiming and Yi, Chenyu and Zhang, Qixin and Shen, Bingquan and Kot, Alex C and Jiang, Xudong},
  booktitle=cvpr,
  year={2026}
}

@inproceedings{zholus2025tapnext,
  title={Tapnext: Tracking any point (tap) as next token prediction},
  author={Zholus, Artem and Doersch, Carl and Yang, Yi and Koppula, Skanda and Patraucean, Viorica and He, Xu Owen and Rocco, Ignacio and Sajjadi, Mehdi SM and Chandar, Sarath and Goroshin, Ross},
  booktitle=iccv,
  year={2025}

}

@inproceedings{edstedt2024roma,
  title={Roma: Robust dense feature matching},
  author={Edstedt, Johan and Sun, Qiyu and B{\"o}kman, Georg and Wadenb{\"a}ck, M{\aa}rten and Felsberg, Michael},
  booktitle=cvpr,
  year={2024}
}

@inproceedings{liu2024grounding,
  title={Grounding dino: Marrying dino with grounded pre-training for open-set object detection},
  author={Liu, Shilong and Zeng, Zhaoyang and Ren, Tianhe and Li, Feng and Zhang, Hao and Yang, Jie and Jiang, Qing and Li, Chunyuan and Yang, Jianwei and Su, Hang and others},
  booktitle=eccv,
  year={2024}
}

@inproceedings{ravi2025sam,
  title={Sam 2: Segment anything in images and videos},
  author={Ravi, Nikhila and Gabeur, Valentin and Hu, Yuan-Ting and Hu, Ronghang and Ryali, Chaitanya and Ma, Tengyu and Khedr, Haitham and R{\"a}dle, Roman and Rolland, Chloe and Gustafson, Laura and others},
  booktitle=ICLR,
  year={2025}
}

@InProceedings{Jung_2026_CVPR,
    author    = {Jung, Sebastian and Zholus, Artem and Sundermeyer, Martin and Doersch, Carl and Goroshin, Ross and Tan, David Joseph and Chandar, Sarath and Triebel, Rudolph and Tombari, Federico},
    title     = {TAPNext++: What's Next for Tracking Any Point (TAP)?},
    booktitle = {Proceedings of the IEEE/CVF Conference on Computer Vision and Pattern Recognition (CVPR) Findings},
    year      = {2026}
}

@article{fischler1981random,
  title={Random sample consensus: a paradigm for model fitting with applications to image analysis and automated cartography},
  author={Fischler, Martin A and Bolles, Robert C},
  journal={Communications of the ACM},
  year={1981},
}

@inproceedings{
zhang2026dila,
title={Di{LA}: Disentangled Latent Action World Models},
author={Tianqiu Zhang and Muyang Lyu and Yufan Zhang and Fang Fang and Si Wu},
booktitle=icml,
year={2026},
}

@inproceedings{
zhang2026structure,
title={Structure Abstraction and Generalization in a Hippocampal-Entorhinal Inspired World Model},
author={Tianqiu Zhang and Muyang Lyu and Xiao Liu and Si Wu},
booktitle=icml,
year={2026},
}

@inproceedings{
fu2022patchfool,
title={Patch-Fool: Are Vision Transformers Always Robust Against Adversarial Perturbations?},
author={Yonggan Fu and Shunyao Zhang and Shang Wu and Cheng Wan and Yingyan Lin},
booktitle=ICLR,
year={2022},
}

@inproceedings{zhang2025phycamo,
  title={Phycamo: A robust physical camouflage via contrastive learning for multi-view physical adversarial attack},
  author={Zhang, Ximin and Chen, Jinyin and Zheng, Haibin and Liu, Zhenguang},
  booktitle=AAAI,
  year={2025}
}

@inproceedings{kim2023breaking,
  title={Breaking temporal consistency: Generating video universal adversarial perturbations using image models},
  author={Kim, Hee-Seon and Son, Minji and Kim, Minbeom and Kwon, Myung-Joon and Kim, Changick},
  booktitle=iccv,
  year={2023}
}

@inproceedings{mur2026v,
  title={V-jepa 2.1: Unlocking dense features in video self-supervised learning},
  author={Mur-Labadia, Lorenzo and Muckley, Matthew and Bar, Amir and Assran, Mido and Sinha, Koustuv and Rabbat, Mike and LeCun, Yann and Ballas, Nicolas},
  booktitle=eccv,
  year={2026},
}

@article{wang2026partially,
  title={Partially Observable Adversarial Patch Attacks on Vision-Language-Action Models in Robotics},
  author={Wang, Xiaofei and Han, Mingliang and Hao, Tianyu and Yang, Yi and Zhao, Yun-Bo and Tang, Keke},
  journal={IEEE Robotics and Automation Letters},
  year={2026},
}

@inproceedings{lou20253d,
  title={3d gaussian splatting driven multi-view robust physical adversarial camouflage generation},
  author={Lou, Tianrui and Jia, Xiaojun and Liang, Siyuan and Liang, Jiawei and Zhang, Ming and Xiao, Yanjun and Cao, Xiaochun},
  booktitle=iccv,
  year={2025}
}
\bibliographystyle{iclr2027_conference}

\clearpage
\appendix
\section{Appendix}

\subsection{Trajectory-Aware Weighting}
\label{app:trajectory_weighting}

To balance the contributions of video--trajectory instances with different levels of texture exposure, we introduce trajectory-aware weighting during task refinement.
Let $\alpha_{itv}(p)\in[0,1]$ denote the visible texture alpha at pixel $p$ in frame $t$ and inference view $v$, obtained from the frame-wise homography and visibility mask.
The total observable texture area of instance $i$ is $a_i=\sum_{v,t,p}\alpha_{itv}(p)>0$, with a single view for models that do not use multi-view inference.
For each minibatch ${B}$, we set $a_{\mathrm{ref}}=\operatorname{median}_{j\in\mathcal{B}}a_j$ and compute the instance weights as:
\begin{equation}
\tilde{w}_i=\operatorname{clip}\!\left(\sqrt{\frac{a_{\mathrm{ref}}}{a_i}},\,0.5,\,2\right),\qquad
w_i=\frac{\tilde{w}_i}{\sum_{j\in{B}}\tilde{w}_j}.
\label{eq:trajectory_aware_refinement_weight}
\end{equation}
This scheme increases the relative contribution of instances with smaller observable texture areas, while clipping limits extreme weights and normalization ensures that $\sum_{i\in\mathcal{B}}w_i=1$.
As the observation trajectories are fixed, the weights are computed independently of the adversarial texture and remain constant during optimization.

\subsection{Attack details and visualization for DILA generation task}
\label{app: attack for dila}
\begin{figure*}[h]
\centering
\includegraphics[width=\textwidth]{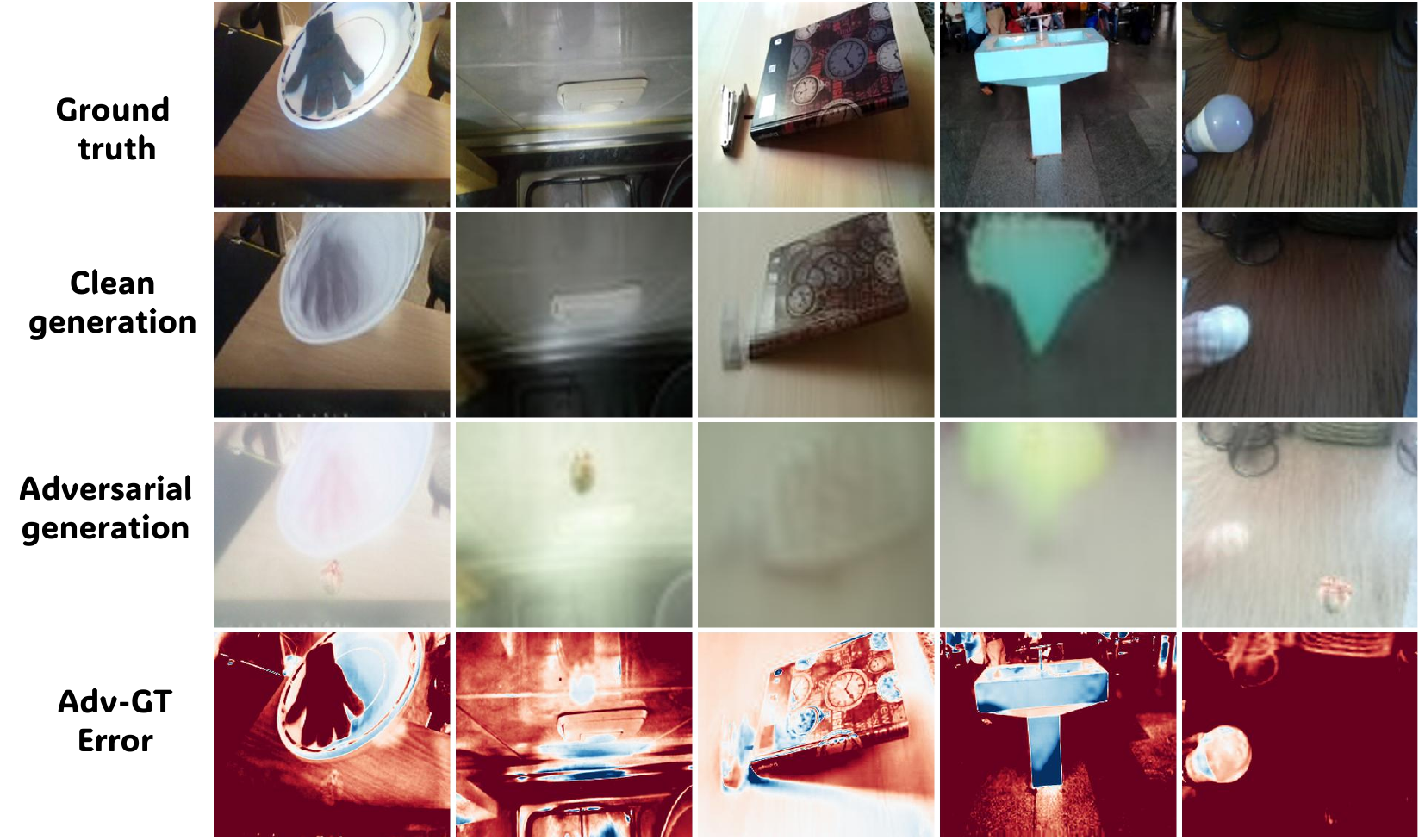}
\caption{Visualization of DiLA reconstruction results under TARA. Each column is a different validation clip.}
\label{fig:dila_qualitative}
\end{figure*}

\textbf{Attack details.}
For DiLA reconstruction, we use 155 Train and 82 Val clips, with 16 attacked source frames.
Contrastive alignment optimizes the mean future spatiotemporal embedding for 10 epochs using Adam ($0.03\to0.003$).
The subsequent five-epoch refinement uses Adam ($0.03\to0.005$) to maximize outside-patch RGB MSE relative to the clean reconstruction.
Both stages use a logical batch of 32 clips and 24 updates per epoch; latent actions are recomputed from attacked source frames.

\textbf{Visualization of the attacked generation.}
Figure~\ref{fig:dila_qualitative} shows five validation clips selected for high MSE gain outside the patch region.
The first row shows the ground truth observed frame, followed by clean reconstruction, adversarial reconstruction, and the change in per-pixel squared error.
The adversarial reconstructions exhibit appearance changes well beyond the attached texture, demonstrating that our attack remains effective at disrupting conditional reconstruction.
\subsection{Generalization Across Content and Trajectories}
\label{app:content-trajectory-generalization}
\begin{table}[h]
\centering
\vspace{-1mm}
\caption{Generalization across video content and observation trajectories on EK100 with V-JEPA2-L. Top-1 ASR (\%, $\uparrow$) is reported on 74 eligible clip instances per condition.}
\label{tab:ek100-content-trajectory-generalization}
\footnotesize
\setlength{\tabcolsep}{4pt}
\renewcommand{\arraystretch}{1.15}
\begin{tabular*}{\columnwidth}{@{\extracolsep{\fill}}lcccc@{}}
\toprule
& \multicolumn{2}{c}{Seen Content} & \multicolumn{2}{c}{Unseen Content} \\
\cmidrule(lr){2-3}\cmidrule(l){4-5}
Method & Seen Traj. & Unseen Traj. & Seen Traj. & Unseen Traj. \\
\midrule
Static center & 17.6 & 17.6 & 13.5 & 9.5 \\
EOT & 32.4 & 24.3 & 32.4 & 16.2 \\
PhyCamo & 17.6 & 20.3 & 14.9 & 12.2 \\
Patch-Fool & 12.2 & 13.5 & 10.8 & 10.8 \\
UPA-RFAS & 13.5 & 13.5 & 9.5 & 5.4 \\
BTC-UAP & 21.6 & 20.3 & 18.9 & 13.5 \\
\midrule
\textbf{TARA} & \textbf{40.5} & \textbf{31.5}
              & \textbf{43.2} & \textbf{28.8} \\
\bottomrule
\end{tabular*}
\vspace{-2mm}
\end{table}

\textbf{Experimental setup.}
We evaluate frozen adversarial textures under four combinations of video content and observation trajectories.
The seen-content cohort contains 74 EK100 training clips; the unseen-content cohort contains 74 clip instances from 35 independently captured practical recordings.
For crossed conditions, we remap the homographies to the V-JEPA2-L input viewport while retaining temporal visibility.
The static-center texture is also rendered along each target trajectory, which is trained without trajectory-aware optimization as a baseline reference.

\textbf{Results.}
Table~\ref{tab:ek100-content-trajectory-generalization} shows that TARA achieves the highest ASR in all four conditions.
On seen content with unseen trajectories, TARA reaches 31.5\% ASR, compared with 24.3\% for the strongest baseline, EOT.
When both content and trajectory are unseen, TARA achieves 28.8\%, exceeding EOT by 12.6 percentage points and static center by 19.3 points.
This demonstrates the effectiveness of TARA texture under the joint shift in content and trajectory.

\subsection{Statistical Stability}
\label{app:seed_stability}

\begin{table}[h]
\centering
\caption{Random-seed stability of TARA. Final-epoch Top-1 ASR (\%).
Mean and sample SD are computed across all three seeds; seed 0 is also
reported in the main results.}
\label{tab:tara_seed_stability}
\footnotesize
\setlength{\tabcolsep}{3pt}
\renewcommand{\arraystretch}{1.12}
\begin{tabular}{llccc}
\toprule
Dataset / Model & Seed & Train & Val & Overall \\
\midrule
EK100 / V-JEPA2-L
  & 0 (reported) & 37.7 & 26.1 & 32.1 \\
  & 1 & 37.9 & 26.1 & 32.2 \\
  & 2 & 36.8 & 24.6 & 31.0 \\
& Mean $\pm$ SD & $37.5\pm0.6$ & $25.6\pm0.9$ & $31.8\pm0.7$ \\
\midrule
SSV2 / V-JEPA2-L
  & 0 (reported) & 38.6 & 34.7 & 36.9 \\
  & 1 & 38.4 & 34.0 & 36.5 \\
  & 2 & 37.0 & 34.0 & 35.7 \\
& Mean $\pm$ SD & $38.0\pm0.9$ & $34.2\pm0.4$ & $36.4\pm0.6$ \\
\bottomrule
\end{tabular}
\end{table}

\textbf{Randomness test.} 
To assess the sensitivity of TARA to random initialization, we repeat the experiments with three random seeds on EK100 and SSV2 using V-JEPA2-L. 
We keep the model, data split, attack objective, and training schedule fixed, and vary the random sampling of training batches and trajectories.
As shown in Table~\ref{tab:tara_seed_stability}, TARA exhibits consistent performance across seeds, with small standard deviations for both Train and Val splits. 
The resulting Overall ASR varies only moderately across runs, indicating that the reported performance is robust to random-seed variation.

\subsection{More visualization}
\label{app:more visual}
\begin{figure*}[h]
\centering
\includegraphics[width=0.95\textwidth]{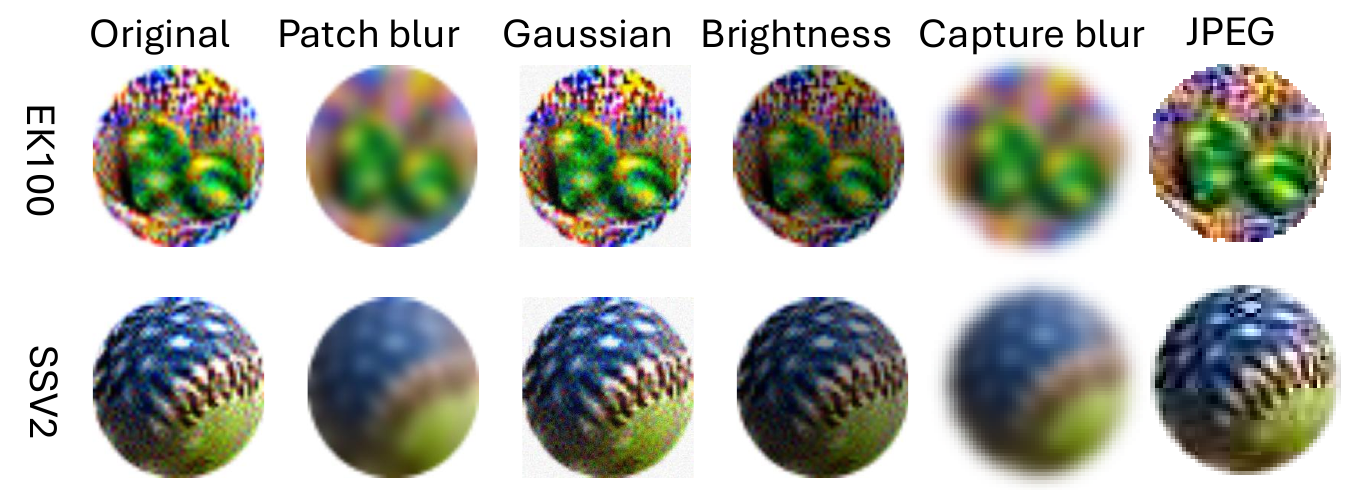}
\caption{Visualization of adversarial textures under different imaging distortions on EK100 and SSV2. Patch and capture blur substantially suppress fine-grained texture details.}
\label{fig:distorted_texture}
\end{figure*}
\textbf{Visualization under simulated capture-and-print distortions.}
Figure~\ref{fig:distorted_texture} compares the original adversarial textures with their distorted counterparts on EK100 and SSV2.
Patch blur and capture blur visibly smooth local texture structures and suppress fine-grained details, providing a qualitative illustration of the performance degradation observed under severe blur.
The remaining distortions alter texture appearance without comparably suppressing its local structures.

\begin{figure*}[h]
\centering
\includegraphics[width=1\textwidth]{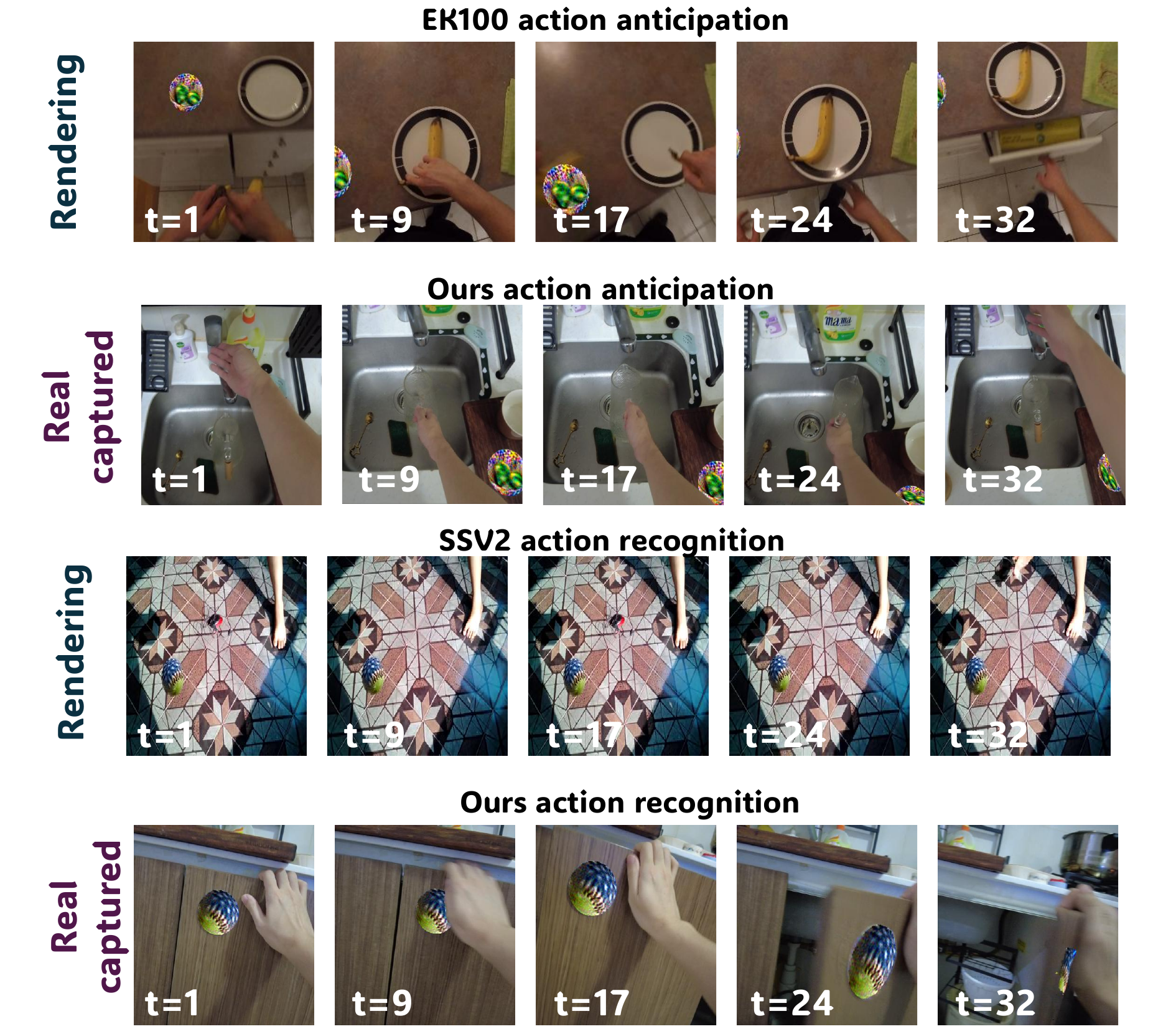}
\caption{Additional trajectory visualizations across digital and physical settings. The first and third rows show digitally projected textures on EK100 and SSV2, respectively, while the second and fourth rows show real-world print-and-recapture recordings with physically attached textures.}
\label{fig:more_traj}
\end{figure*}
\textbf{Additional trajectory visualizations.}
Figure~\ref{fig:more_traj} presents representative examples from two evaluation settings: digitally projected adversarial textures on EK100 and SSV2 videos, and real-world print-and-recapture recordings in which the textures are physically printed and attached to objects.
For each video, we show sampled frames along its observation trajectory to illustrate how a fixed texture is observed under evolving object--camera motion.
Across different video instances, the texture undergoes substantial changes in image location, apparent scale, and visibility, including partial occlusion and movement toward image boundaries.
These examples illustrate the diverse, temporally varying observation conditions considered in both our digital benchmark and physical evaluation.

\subsection{Implementations details}
\label{app:implementation_details}
\begin{table*}[t]
\centering
\vspace{-1mm}
\caption{Model-specific TARA optimization schedules and update budgets.}
\label{tab:tara_implementation_schedules}
\footnotesize
\setlength{\tabcolsep}{4pt}
\renewcommand{\arraystretch}{1.15}
\begin{tabularx}{\textwidth}{@{}p{2.5cm}cp{2.9cm}>{\raggedright\arraybackslash}Xr@{}}
\toprule
Model / task & Train / Val & Alignment feature & Optimization schedule & Updates \\
\midrule
V-JEPA2-L / EK100 & 356 / 249 & Action query
& F10 Adam ($0.03\!\to\!0.003$)\newline FT5 sign-PGD ($0.01\!\to\!0.005$, step) & 360 \\
V-JEPA2.1-B / EK100 & 175 / 140 & Action + token mean
& F10 Adam ($0.03\!\to\!0.003$)\newline FT5 sign-PGD ($0.01\!\to\!0.005$, cosine) & 360 \\
AVT-B / EK100 & 85 / 60 & Future aggregate
& F10 Adam ($0.03\!\to\!0.003$, cosine)\newline FT5 sign-PGD ($0.005\!\to\!0.001$, cosine) & 360 \\
\midrule
V-JEPA2-L / SSV2 & 184 / 144 & Probe prelinear
& F10 Adam ($0.03\!\to\!0.003$)\newline FT5 sign-PGD ($0.01\!\to\!0.005$, step) & 360 \\
VideoMAE-B / SSV2 & 184 / 144 & Classifier input $F_0$
& F10 Adam ($0.03\!\to\!0.003$)\newline FT5 sign-PGD ($0.03\!\to\!0.005$, cosine) & 360 \\
InternVideo2-dist-L / SSV2 & 184 / 144 & Layer-24 token mean
& F10 Adam ($0.03\!\to\!0.003$)\newline FT5 sign-PGD ($0.005\!\to\!0.001$, cosine) & 360 \\
\bottomrule
\end{tabularx}
\vspace{-2mm}
\end{table*}

\textbf{Model-specific attack configurations.}
For EK100 anticipation, V-JEPA2-L uses a 32-frame, $256\times256$ context and aligns its action-query representation.
V-JEPA2.1-B uses a 32-frame, $384\times384$ mixed-window input and aligns its pooled action representation concatenated with the final-token mean.
For AVT-B, it uses its native 10-frame input with three crops and horizontal flips, aligning the aggregated future representation.
For SSV2 recognition, V-JEPA2-L aligns the prelinear probe representation using two 16-frame segments and three crops; VideoMAE-B aligns its classifier-input feature $F_0$ using two 16-frame samples and three crops; and InternVideo2-dist-L aligns the mean layer-24 token representation using four eight-frame offsets and three crops.
Task refinement uses each model's final prediction after native view aggregation, rather than optimizing separate predictions for individual views.

\textbf{Training details and checkpoint selection.}
All classification alignment stages use temperature $\tau=0.1$ and Adam optimizer.
Refinement minimizes the ground-truth log-probability of the model's aggregated task prediction using sign-PGD, with texture values projected to $[0,1]$ after each update.
The training details are shown in ~\ref{tab:tara_implementation_schedules}.
For a fair comparison, all compared attacks use the same optimizer, learning-rate schedule, and number of texture updates as TARA under the corresponding model and task setting, while retaining their respective attack objectives.
All models use the final checkpoint of their alignment stage.
%

\textbf{Real-World Digital and Physical Evaluation.}
For the real-world digital comparison, we utilize the same homography and mask to generate the adversarial video for a fair comparison.
For the physical print-and-recapture evaluation, we attach printed EOT and TARA textures to physical objects and compare the resulting adversarial recordings (ADV) against matched recordings with white textures at the same attachment locations (WHITE), under corresponding actions and recording conditions.
%

\textbf{Random seeds and hardware.}
All reported results use seed 0 for texture initialization and batch/trajectory sampling and represent single-seed runs.
Experiments were conducted on four NVIDIA GeForce RTX A6000 GPUs (48\,GiB each), with each model-specific attack job running on one GPU.
Micro-batching and gradient replay are used where necessary without changing the logical batch size or texture-update budgets.

\subsection{Surface-Trajectory Construction and Quality Control}
\label{app:surface_trajectory}

\textbf{Trajectory construction.}
For each newly annotated clip, an annotator selects a locally planar object surface and a feasible texture attachment region in a reference frame.
We sample anchor points within this region and track them across frames using TAPNext++~\citep{Jung_2026_CVPR,zholus2025tapnext}.
For frame $t$ of clip $i$, we estimate a reference-to-frame homography $\bm G_{it}$ from the tracked correspondences using RANSAC~\citep{fischler1981random}.
The texture-to-frame homography is $\bm H_{it}=\bm G_{it}\bm A_i$, where $\bm A_i$ maps the canonical texture to the reference frame.
Candidate projections are validated using inlier support, held-out reprojection error, spatial coverage, and temporal scale consistency.
For spatial validation, the bounding box of the source fitting points is divided into four quadrants, of which at least three must contain accepted inliers.
When sparse tracking is insufficient, RoMa~\citep{edstedt2024roma} provides dense correspondences with nearby validated frames for geometric recovery.
Table~\ref{tab:trajectory_implementation} summarizes the geometry-estimation and screening settings used for newly annotated trajectories.
GroundingDINO~\citep{liu2024grounding} and SAM2~\citep{ravi2025sam} provide candidate supporting-surface and foreground-occluder masks for estimating visible texture support.

\begin{table}[t]
\centering
\vspace{-1mm}
\caption{Geometry-estimation settings for newly annotated trajectories.}
\label{tab:trajectory_implementation}
\footnotesize
\setlength{\tabcolsep}{3pt}
\renewcommand{\arraystretch}{1.12}
\begin{tabular*}{\columnwidth}{@{\extracolsep{\fill}}ll@{}}
\toprule
Component & Setting \\
\midrule
Reference-point sampling & $12\times12$ grid; 2-pixel erosion \\
Minimum correspondences & $\geq20$ fitting; $\geq8$ held-out \\
RANSAC & 2-pixel threshold; 0.999 confidence; 10,000 iterations \\
Inlier screening & $\geq20$ inliers; ratio $\geq0.60$ \\
Spatial screening & Inliers in $\geq3$ local quadrants \\
Temporal screening & Projection-area continuity \\
\bottomrule
\end{tabular*}
\vspace{-2mm}
\end{table}

\begin{figure*}[t]
\centering
\includegraphics[width=\textwidth]{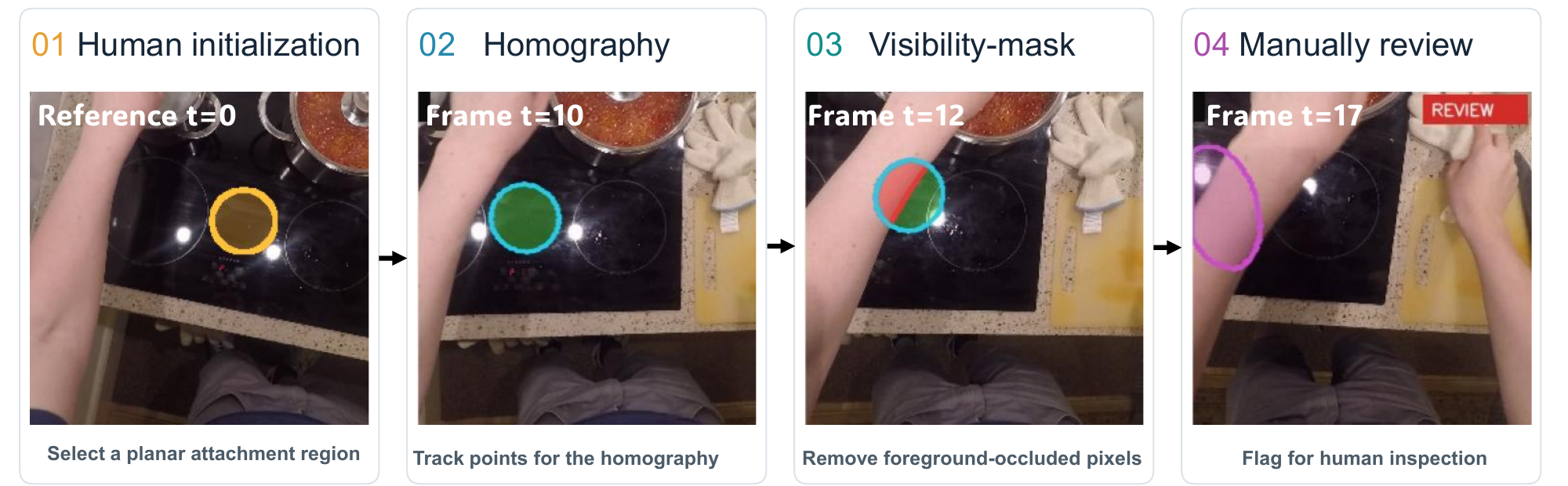}
\caption{Human-guided surface-trajectory construction in an EK100 clip. (1) Manual attachment-region selection. (2) Homography estimation from tracked points. (3) Visible-support estimation under foreground occlusion. (4) Flagging of an inconsistent projection for human review.}
\label{fig:surface_trajectory_qa}
\end{figure*}

\textbf{Human review and quality control.}
Human review guides the initial texture placement and the correction of uncertain geometric and visibility annotations.
Reviewers inspect temporal overlays for surface mismatch, projection drift, abrupt scale changes, and incorrect occlusion.
As illustrated in Figure~\ref{fig:surface_trajectory_qa}, they may revise the fixed attachment region for an entire clip, reselect anchor points and rerun tracking, verify a recovered projection, or correct its visible support.
Tracking failure on a still-visible surface triggers geometric review rather than an automatic assignment of zero visibility.
Zero visibility is assigned only when the texture is confirmed to be fully occluded or outside the image.
Unresolved frames retain an explicit unknown status rather than being treated as physically invisible.

\textbf{Benchmark composition and rendering.}
The digital benchmark contains 605 EK100 clips (356 Train/249 Val) and 328 SSV2 clips (184 Train/144 Val), with trajectory records for 19,251 and 10,204 distinct raw frames, respectively.
These counts include frames with unresolved geometry.
SSV2 clips were selected based on V-JEPA2-L clean Top-1 correctness and geometric eligibility; attack success is computed separately using each evaluated model's clean-correct examples.
Clip identities, splits, and trajectory annotations are fixed across attack methods, and only Train clips contribute texture gradients.
The stored frame-wise annotations include a validity status in addition to the homography and visibility mask defining $\gamma_i$.
We render validated projections using their recorded geometry and visibility; confirmed invisibility and unresolved geometry both leave the input frame unchanged.
The latter remains labeled unknown rather than physically invisible, while valid projections in other frames of the same clip are retained.
For each evaluated model, the annotations are aligned with its sampled raw frames and transformed consistently with its resizing and cropping operations.

\textbf{Statistical definitions.}
The digital statistics in Table~\ref{tab:trajectory_stats} use a fixed sampling and center-crop configuration for each dataset, whereas annotated-frame counts refer to distinct raw frames with trajectory records.
Homography coverage is the fraction of sampled positions with a validated homography and a nonempty projected texture region within the crop.
Mask rate is the fraction of projected positions with assessable foreground annotations for which an occluder overlaps the projected region; unknown occlusion states are excluded.
Texture area measures the pooled fraction of image pixels covered by the recorded visible texture support:
\begin{equation}
\mathrm{TextureArea}
=100\frac{\sum_{(i,t)\in\mathcal{S}}|\bm M_{it}|}
{\sum_{(i,t)\in\mathcal{S}}|\bm I_{it}|},
\label{eq:texture_area}
\end{equation}
where $\mathcal{S}$ denotes all sampled positions and $\bm I_{it}$ is the corresponding model input crop.
Unresolved positions contribute zero rendered area but remain in the denominator; thus, this metric measures annotation-supported rendering coverage rather than independently verified physical visibility.
Texture motion is the median across clips of the maximum pairwise projected-center displacement normalized by the crop diagonal, computed for clips with at least two distinct valid raw-frame projections.
Figure~\ref{fig:trajectory_statistics} additionally reports frame-level visible fraction, $|\bm M_{it}|/|\bm P_{it}|$, for resolved visibility annotations, and projected area, $100|\bm P_{it}|/|\bm I_{it}|$, for valid nonempty projections, where $\bm P_{it}$ denotes the projected texture region before foreground removal.

\subsection{Examples of rendered and real-world captured clips.}
\label{sec:full traj}
\begin{figure*}[t]
\centering
\includegraphics[width=\textwidth]{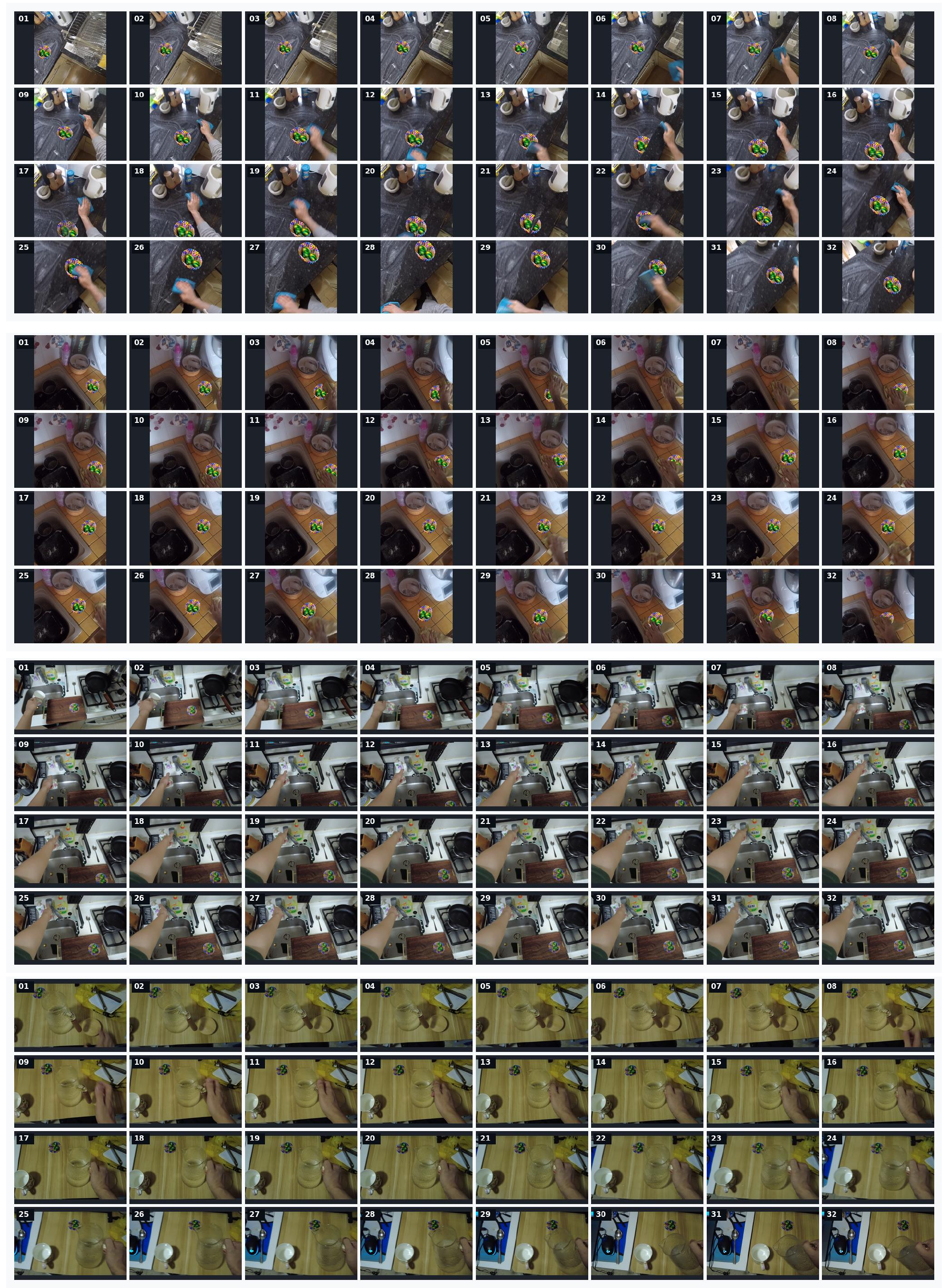}
\caption{Visualization of the clips in action anticipation. The first two samples are rendered from the EK100 dataset, and the last two samples are our real-world print-and-capture data.}
\label{fig:full_trajectory_anti}
\end{figure*}

\begin{figure*}[t]
\centering
\includegraphics[width=\textwidth]{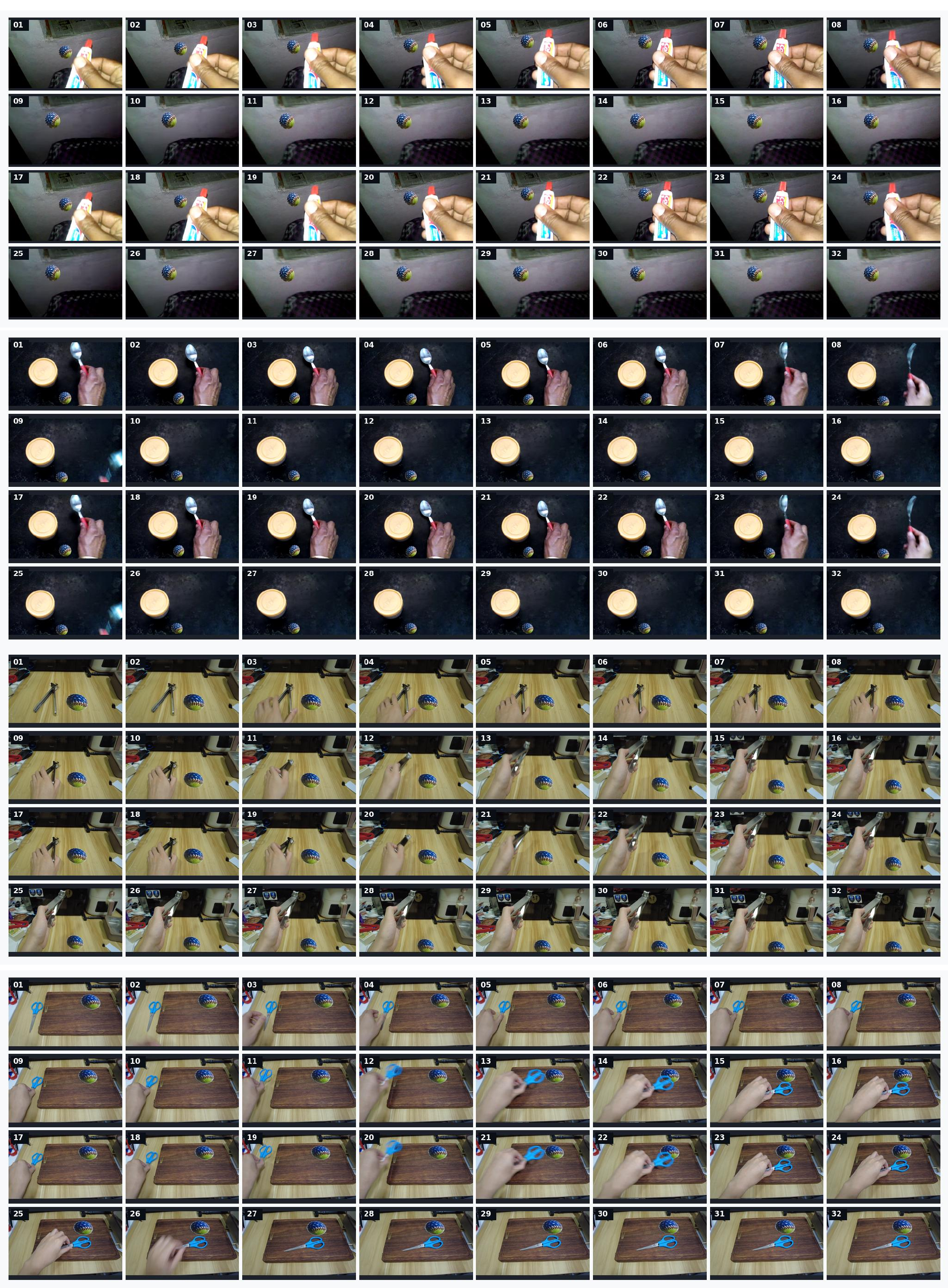}
\caption{Visualization of the clips in action recognition. The first two samples are rendered from the SSV2 dataset, and the last two samples are our real-world print-and-capture data.}
\label{fig:full_trajectory_reco}
\end{figure*}

Figures~\ref{fig:full_trajectory_anti} and~\ref{fig:full_trajectory_reco} show complete 32-frame model inputs for action anticipation and recognition, respectively.
In each figure, the first two clips use textures digitally rendered along annotated trajectories, while the last two show physically printed textures captured in real-world videos.
These examples illustrate how the same surface-attached texture is observed across changing viewpoints, scales, and occlusions.

\end{document}